%% file: acl2023.tex
\documentclass[11pt]{article}

\usepackage{acl2023}
\usepackage{times}
\usepackage{latexsym}
\usepackage[T1]{fontenc}
\usepackage[utf8]{inputenc}
\usepackage{microtype}
\usepackage{graphicx}
\usepackage{booktabs}
\usepackage{amsmath}
\usepackage{amssymb}
\usepackage{bm}
\usepackage{multirow}
\usepackage{makecell}
\usepackage{xcolor}
\usepackage{colortbl}
\usepackage{url}
\usepackage{xspace}
\usepackage{pifont}
\usepackage{fontawesome}
\usepackage{wrapfig}
\usepackage[ruled,vlined,linesnumbered]{algorithm2e}
\usepackage{tikz}
\usetikzlibrary{arrows.meta,positioning,calc,fit,shapes.geometric}

\newcommand{\method}{\textsc{WebWorld}\xspace}

\newcommand{\iterstep}{\textsc{IterStep}\xspace}
\newcommand{\qualstep}{\textsc{QualityStep}\xspace}

\definecolor{HeaderColor}{RGB}{226,230,236}
\definecolor{GoodColor}{RGB}{248,250,253}
\definecolor{ClosedColor}{RGB}{248,242,228}
\definecolor{FairColor}{RGB}{248,248,248}
\definecolor{OurColor}{RGB}{215,234,222}
\definecolor{tablegray}{gray}{0.92}

\title{WebWorld: The Browser as a World Model for Self-Improving Web Code}
\ifdefined\aclreviewcopy
  \author{Anonymous ACL submission}
\else
  \author{%
    {\bf Jiajun Wu}\textsuperscript{\rm 1},
    {\bf Jian Yang}\textsuperscript{\rm 1,$\dagger$},
    {\bf Yaxin Du}\textsuperscript{\rm 3},
    {\bf Wei Zhang}\textsuperscript{\rm 1},
    {\bf Haowen Wang}\textsuperscript{\rm 2},
    \\
    {\bf Junhang Cheng}\textsuperscript{\rm 1},
    {\bf Yuxuan Zhang}\textsuperscript{\rm 2},
    {\bf Tuney Zheng}\textsuperscript{\rm 2},
    {\bf Xianglong Liu}\textsuperscript{\rm 1},
    {\bf Ming Zhou}\textsuperscript{\rm 4}
    \\
    \textsuperscript{\rm 1}\,Beihang University;
    \textsuperscript{\rm 2}\,IQuest Research;
    \textsuperscript{\rm 3}\,Shanghai Jiao Tong University
    \\
    \textsuperscript{\rm 4}\,Langboat
    \\
    $\dagger$Corresponding Authors. Email: \texttt{jiayang@buaa.edu.cn}
  }
\fi

\begin{document}
\raggedbottom
\maketitle

\begin{abstract}
VLM-driven self-improvement of web code has a structural flaw: the model that proposes the repair is the model that judges it, and visual plausibility under that judge is a poor proxy for whether the page actually works. What the loop is missing is a counterparty the VLM cannot fool, and the browser already is that counterparty: a deterministic, executable simulator of how an HTML artifact behaves under user actions, and in everything but name a world model for web code. We present \method, the interface that lets a VLM prior interact with this browser-as-world-model autonomously and decides which interactions become supervision. Each round, the VLM emits a critique that the planner compiles into a typed interaction contract; the browser re-executes the candidate and issues an acceptance certificate only when both target progress and preservation of every previously verified capability hold; certified transitions accumulate as a quality ratchet that is the only thing the SFT export ever sees. Under matched training, \method-27B improves Raw-27B by 5.3 points on HTMLBench-400 and 14.9 points on MiniAppBench-Val, and reaches the level of strong frontier systems such as Kimi-K2.6 and GPT-5.4 on interactive HTML generation. Equal-size ablations show that browser-backed admission carries the gain: without the certificate, the matched 9B lift nearly disappears.
\end{abstract}

\section{Introduction}
\label{sec:intro}

LLMs and VLMs can now produce complete front-end artifacts from a short prompt, but a page that looks right at first render is not the same page that works. Submit buttons can be inert, games can ignore keyboard input, and a repair can polish one region while silently removing a previously working control. The natural response, closing the loop with a VLM critique-and-rewrite cycle, does not by itself solve this. In our experiments (\autoref{sec:experiments}), the closest re-implementation of that naive loop, which keeps VLM critique and skill routing but drops the browser-issued certificate, lifts HTMLBench by only \textbf{0.4} points over Raw, more than ten times less than the full \method gate. Most of what the loop calls progress is noise unless something outside the VLM admits it.

The reason is structural. The VLM is doing two jobs at once: it is the proposer, broad in coverage and useful at suggesting plausible repairs, and also the judge that decides which repair to keep. Both judgments live in screenshot space, so a winning candidate is one that wins visually, exactly the wrong criterion when the failures are behavioral. Better prompting cannot close this gap, because closure has to come from evidence the VLM did not produce. The proposer needs a counterparty it cannot fool: a browser that re-executes the artifact.

\begin{figure}[t]
\centering
\includegraphics[width=0.96\columnwidth]{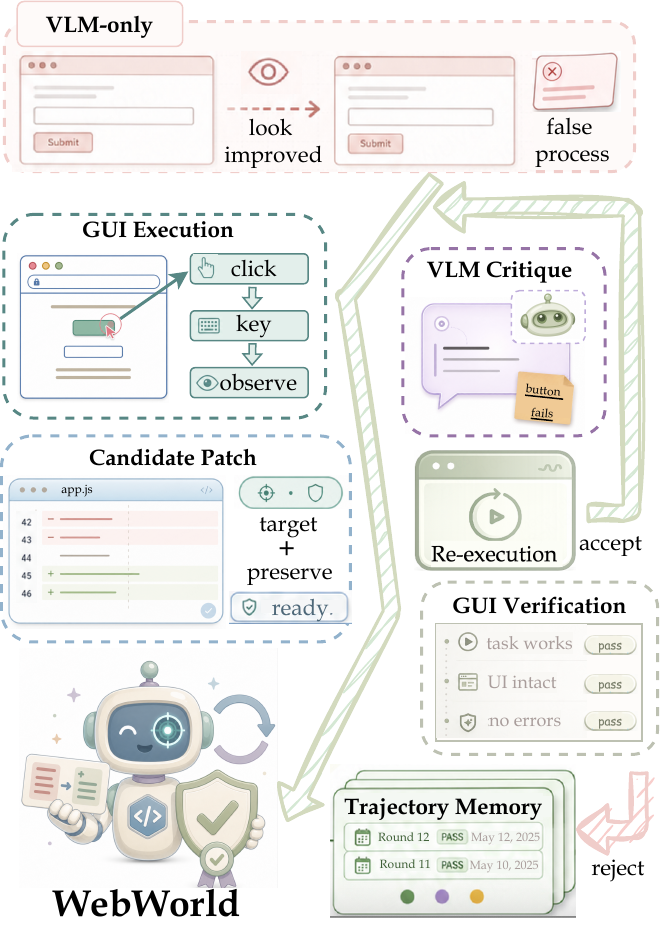}
\caption{When the VLM judges by screenshot, edits that look improved can fail under browser re-execution: the proposer-judge gap our audit measures.}
\label{fig:intro}
\vspace{-10pt}
\end{figure}

The browser is that counterparty, already built. It is a deterministic, executable simulator of how an HTML artifact behaves under user actions, and in everything but name a world model for web code (\autoref{fig:intro}). We do not have to train one, which is what makes this problem unusually clean: the world model is fully specified by the platform, and the open question is the interface that lets a VLM prior interact with it autonomously and decides which interactions become supervision.

\begin{figure*}[t]
\centering
\includegraphics[width=0.98\textwidth]{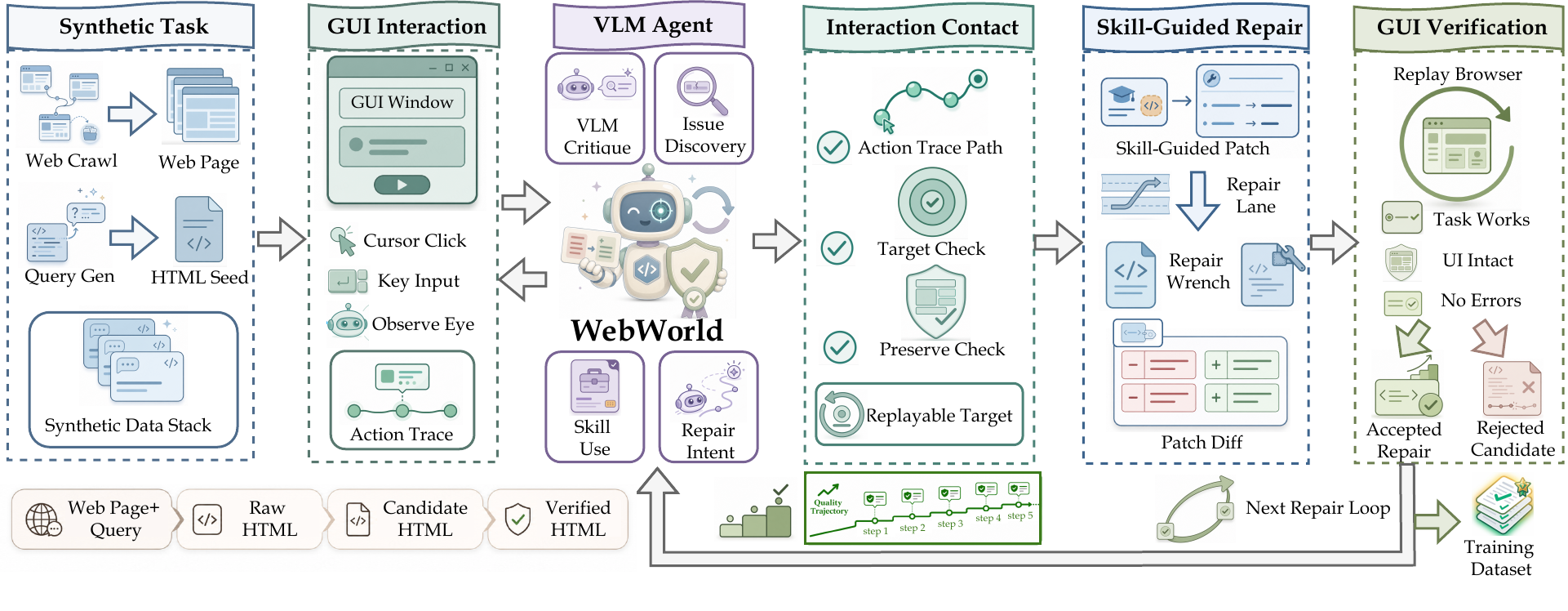}
\caption{\method splits hypothesis from proof: the VLM proposes via an interaction contract, the browser re-executes to issue or refuse an acceptance certificate, and only certified transitions enter the SFT export.}
\label{fig:framework}
\vspace{-15pt}
\end{figure*}

We present \method as this interface, sketched in \autoref{fig:framework}. At each round, the runtime executes the artifact and records GUI evidence. The VLM proposes a critique; the planner compiles it into a typed interaction contract with a target predicate, replay action, preserve set, and evidence path. The browser re-executes the candidate and issues an acceptance certificate only when target progress holds and every previously verified capability is preserved. Certified transitions form a quality ratchet and the SFT export. Hypothesis and proof are split by construction: the VLM proposes, and only the browser world model admits, so that failure cannot contaminate supervision.

\textbf{(1) Browser as a ready-made world model for web code.} We re-frame autonomous HTML improvement as a world-model agent problem and resolve it without training a world model. The browser already plays that role, and the unit of supervision becomes a certified state transition rather than a final screenshot or scalar preference.

\textbf{(2) Contract, certificate, and ratchet.} Three coupled artifacts let the VLM prior and browser world model communicate: a typed interaction contract that constrains repair, an acceptance certificate issued only when re-execution proves target progress and preservation, and a quality ratchet that admits certified transitions to the next baseline and training pool.

\textbf{(3) Evidence from browser proof.} The results show the world model is doing the work: \method-27B reaches \textbf{52.7} on HTMLBench-400 and \textbf{85.5} on MiniAppBench-Val, improving over Raw-27B by \textbf{5.3} and \textbf{14.9} points while matching frontier systems such as Kimi-K2.6 and GPT-5.4 on interactive HTML generation. 9B ablations show re-execution is essential: without the certificate, the gain over Raw is only $0.4$ point. The browser is the certifying world model, not a passive viewer.

\section{\method}
\label{sec:method}

\subsection{Overview}
\label{sec:method_overview}

A naive critique-and-rewrite loop shows the VLM a screenshot, asks for an improvement, and trains on whatever comes back. Two failure modes follow: the post-edit screenshot can look better while the page is in fact broken under interaction, and an isolated improvement can silently regress behavior that an earlier round had carefully established. Closing the loop autonomously means giving the VLM something stricter to argue with than its own screenshot preference.

The natural counterparty is the browser itself. The browser already simulates how an HTML artifact behaves under user actions, deterministically and at scale; in everything but name it is a world model for web code. The VLM supplies the missing half, a learned prior over plausible repairs, but it is unreliable as the judge of its own output, so it cannot be trusted to close the loop on its own. \method is the interface that lets the prior interact with this browser-as-world-model autonomously and decides which interactions become supervision. No human intervenes; the contract planner and the certificate gate are deterministic policies over typed inputs.

The interface is organized around three coupled artifacts, each answering a specific failure of the naive loop. The interaction contract (\S\ref{sec:critique}) compiles a free-form critique into a binding claim the browser can later check, removing the ``looks polished'' judgments that the screenshot-only loop anchors on. The acceptance certificate (\S\ref{sec:verification}) is the only way a candidate becomes the next baseline: the browser re-executes the contract and either issues a proof that the target is reached and prior behavior is preserved, or refuses with a typed rejection. The quality ratchet (\S\ref{sec:ratchet}) admits only certified transitions into both the agent's capability memory and the SFT export, separating runtime exploration from training supervision.

\autoref{fig:framework} threads these artifacts together. At each round the runtime renders the current artifact in a real browser, drives it through a fixed catalogue of interaction probes, and records an observation of screenshots, DOM and console state, probe outcomes, and the executed action trace. The VLM reads that observation and emits a critique that names an issue family and the evidence it rests on; the planner compiles the critique into a contract; the router selects the repair skill matching the issue; the skill proposes a candidate edit; the runtime re-executes the candidate under the contract; and the certificate gate either admits it as the next baseline or refuses with a typed rejection that stays in the routing stream but never reaches the training pool. The probe catalogue and the skill inventory are listed in \autoref{app:implementation}; the method does not depend on the particular list.

Four properties hold by construction throughout the loop. Hypothesis--proof separation: the VLM proposes; only the browser admits, so no claim enters the training pool without re-execution. Typed claims: every critique reduces to predicates over observable browser state, so the system has no perceptual-only judgments. Monotone capability memory: the verified set $\mathcal{P}_t$ grows only by certified evidence, and any later patch that breaks an element of it is rejected at the next round. Deterministic admission: contract admission and certificate gating are functions of typed inputs alone, so every accept and reject is reproducible off-line.

\subsection{World-Model-Verified Trajectories}
\label{sec:trajectory}

We model self-improvement as a sequence of state transitions verified by the browser as the world model. At round $t$, the verifier evaluates a candidate against the current request, repair contract, candidate artifact, and the set of already verified predicates $\mathcal{P}_t$:
\begin{equation}
  \mathcal{V}(q, x_t, r_t, \hat{x}_{t+1}, \mathcal{P}_t) \rightarrow
  \{\textsc{reject}, \textsc{accept}(e_t)\}.
\end{equation}
An accepted trajectory is
\begin{equation}
\begin{aligned}
  \tau = [&(x_0,c_0,r_0,x_1,e_0), \ldots,\\
          &(x_{T-1},c_{T-1},r_{T-1},x_T,e_{T-1})],
\end{aligned}
\end{equation}
where each transition must show target progress and preserve prior capabilities:
\begin{equation}
\begin{aligned}
  \textsc{target}(r_t, x_t)&=0,\\
  \textsc{target}(r_t, x_{t+1})&=1,\\
  p(x_{t+1})&=1,\quad \forall p\in\mathcal{P}_t.
\end{aligned}
\end{equation}
Training uses the transition $(x_t,c_t,r_t,x_{t+1},e_t)$ rather than a final HTML file. The contract $r_t$ fixes what the candidate must achieve and preserve; the certificate $e_t$ records which evidence admitted it.

\subsection{Interaction Contract}
\label{sec:critique}

The first job of the interface is to turn the VLM's free-form judgment into something the browser can check. The VLM reads $o_t$ rather than source code and emits a structured critique: the issue family it sees, the evidence in $o_t$ that supports it, the affected region, the success condition the post-repair page should satisfy, the behavior that must be preserved, and a suggested repair skill. ``Make it more polished'' is not a valid critique; every part must point at evidence the runtime already recorded.

From that critique the planner compiles an interaction contract $r_t$ that the verifier later treats as a binding claim. A contract is a small triple of commitments: what the repair has to achieve, how the browser will check it, and what already-working behavior must survive. The target is a typed predicate over the post-repair page state; the check is a concrete sequence of browser actions, anchored to stable selectors, that the verifier can re-run to decide that predicate; the preservation requirement is the set of predicates inherited from earlier certified rounds. When the defect is local, the contract also pins an impact scope, so edits that migrate to unrelated regions fail the contract itself.

Compiling the critique into a contract is also the system's first filter. A contract is rejected up front if its claim cannot in principle be certified later: critiques anchored to broad containers, perception-only descriptions, or DOM-state predicates with no screenshot-observable consequence never reach the repair stage. Once admitted, it narrows repair: unrelated restyling leaves replay and predicate memory unchanged, giving no certifying evidence.

The candidate edit is generated by a typed repair skill, a constrained patcher whose output diff is bounded by the contract's impact scope, rather than by free-form rewrite. A pre-execution filter discards candidates that cannot move the contract forward (visual-only edits outside the target region, duplicate edits, or patches whose scope intersects the preserve set), so they never reach the browser; the full skill and patch-lane inventory is in \autoref{app:implementation}. This routing is proposer-side machinery: it sharpens the hypothesis before it reaches the world model, but it is not itself a verification step.

\subsection{Acceptance Certificate}
\label{sec:verification}

\input{table_main_results}

The contract is a claim; the certificate is the proof. A candidate is admitted only when re-execution under the contract produces an acceptance certificate $e_t$: a self-contained record that restates the objective, summarizes the diff the candidate applied, names the proof level the verifier used to discharge the contract, and packages the executed evidence (screenshots, action trace, probe outcomes) needed to reproduce the decision off-line. Each accepted candidate carries exactly one certificate, and each certificate is bound to exactly one contract; nothing else can promote a patch into the trajectory. Typed refusals (partial progress, visual-only, preserve-risk, stale replay) make the certificate-rule ablation in \autoref{sec:experiments} a clean gate-component test.

Proof levels are tried in priority order. Same-trace replay comes first: the system replays the contract's exact action. Capability gain is next: a previously failing probe macro now passes. Target-issue progress follows: the target predicate becomes true under an equivalent replay. Localized visual evidence comes after: the patch diff and the visual change both lie inside the contract's impact scope. Static structural repair is last: a non-interactive content addition with low preserve risk. A certificate records only the winning proof level, making later audits replayable rather than subjective.

\subsection{Quality Ratchet}
\label{sec:ratchet}

\method splits runtime progress from training-quality progress, which most critique-and-rewrite loops conflate. An \iterstep is any safe runtime step that keeps exploration moving, such as a re-render, a probe run, a partial fix, or a refused candidate. A \qualstep is a certificate-backed transition that can be exported as supervision. For each candidate repair the runtime constructs
\begin{equation}
  z_t=(q,x_t,o_t,c_t,r_t,\hat{x}_{t+1},e_t,\rho_t),
\end{equation}
where $e_t$ is the certificate, which is empty when refused, and $\rho_t$ is the typed rejection metadata.

The admission rule is stricter than non-regression. It marks a candidate as a \qualstep only when both checks hold:
\begin{equation}
\begin{gathered}
  \textsc{target}(r_t,\hat{x}_{t+1})=1,\\
  p(\hat{x}_{t+1})=1\quad \forall p\in\mathcal{P}_t.
\end{gathered}
\end{equation}
When the rule fires, the runtime accepts $\hat{x}_{t+1}$ as the new baseline and updates memory:
\begin{equation}
  \mathcal{P}_{t+1}=\mathcal{P}_t\cup\textsc{verified}(e_t).
\end{equation}
A later patch that fixes a new target but breaks a previously certified control is then rejected at the next round, instead of silently overwriting prior progress. When the rule does not fire, the candidate stays in the \iterstep stream for routing and debugging but cannot enter the trajectory or the SFT export. Thus the browser judges transitions, not pages: it admits only executable progress under inherited state.

The SFT export is therefore
\begin{equation}
\begin{aligned}
  \mathcal{D}_{\mathrm{SFT}}
  =
  \{&(q,o_t,c_t,r_t,x_{t+1},e_t)\mid\\
    &\qualstep(z_t)=1\}.
\end{aligned}
\end{equation}
The exported target is an artifact paired with the executed failure it addressed and the certificate that admitted it. The appendix case studies in \S\ref{app:rendered_cases} show this ratchet round by round, including a rejected round whose diagnostic feeds back into the next contract rather than the export pool.

\section{Experiments}
\label{sec:experiments}

\subsection{Setup}

We test the \autoref{sec:intro} claim: browser-verified transitions train better behavior than raw HTML supervision under a matched recipe. The \method corpus is $32{,}800$ certificate-accepted transitions; Raw and \method routes are trained at 4B/9B/27B on this budget, and 9B mechanism ablations use the same budget per acceptance rule. Backbone family, prompt template, optimizer, and evaluation runner are fixed across rows; only the supervision source changes. Evaluation uses interactive benchmarks whose executable artifacts are scored by browser behavior: HTMLBench-400~\citep{htmlcure2026} as the primary test and MiniAppBench-Val~\citep{zhang2026miniappbench} for transfer.

\begin{figure}[t]
\centering
\includegraphics[width=\columnwidth]{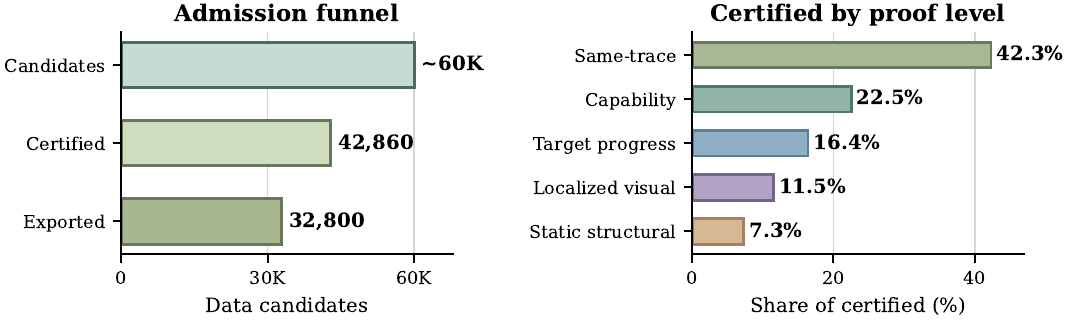}
\caption{Admission funnel and proof-level decomposition of the certified pool.}
\label{fig:funnel}
\vspace{-10pt}
\end{figure}

\autoref{fig:funnel} shows construction as a multi-round critique-repair-verify loop. Each round applies contract admission and certificate issuance, feeds typed rejections into the next critique, and accumulates verified capability with depth. Same-trace replay and capability gain dominate the certified pool; re-execution evidence, not visual preference, makes a transition exportable. Thus the experiment evaluates the interface that selects supervision, not just a larger filtered dataset. Counts and pipeline parameters are in \autoref{app:funnel_details}.

\paragraph{Benchmark independence.}
The construction pipeline never queries benchmark items. \autoref{app:leakage_audit} audits the split. Artifacts, VLM-critique prompts, and runtime probes are disjoint from the 400 prompts and their 6{,}000 deterministic test cases. WebWorld probes use generic actions such as click, drag, keyboard, and form fill; HTMLBench-400 uses item-specific assertions from a separate team. The HTMLCure runner is held-out evaluation only and is never queried during data construction or training.

\subsection{Main SFT Results}

\begin{figure}[t]
\centering
\includegraphics[width=\columnwidth]{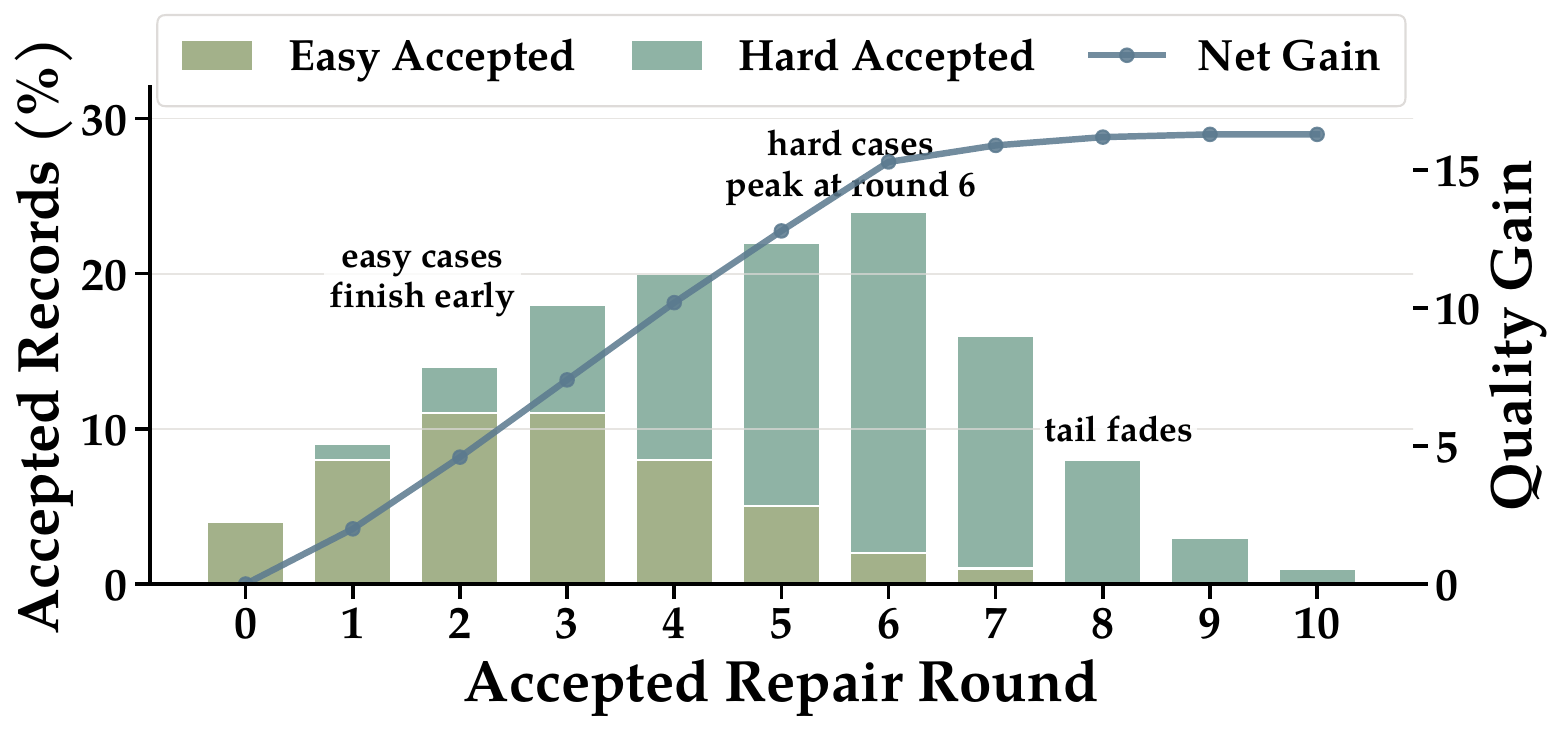}
\caption{Accepted repairs span multiple rounds.}
\label{fig:analysis_iteration_rounds}
\vspace{-10pt}
\end{figure}

The first contribution test asks whether SFT on certificate-accepted transitions produces models that behave better than SFT on raw HTML under a matched recipe. \autoref{tab:main_sft_results} shows it does, and the gap widens with capacity. At 27B, \method reaches $\textbf{52.7}$ on HTMLBench with $\textbf{43.2\%}$ TC pass and $\textbf{85.5}$ on MiniAppBench-Val, improving Raw-27B by $\textbf{5.3}$ score points, $\textbf{9.7}$ TC-pass points, and $\textbf{14.9}$ MiniApp points. As benchmark context rather than a controlled model comparison, this exceeds Kimi-K2.6 ($49.8$) and GPT-5.4 ($49.2$) on HTMLBench, while matching Kimi-K2.6 and staying close to GPT-5.4 on MiniApp. The lift concentrates on the dimensions the certificate gate actually constrains: TC pass and Functionality move strongly with the score at every scale, and MiniApp transfer improves by $10.5$ to $14.9$ points across 4B--27B. Rendering, Visual, and Code differ by less than a point in either direction with no consistent sign across scales; the gate does not optimize for screenshot polish, and it does not have to in order to win on behavior. The gap widens with capacity: $3.4$ points at 4B ($46.7$ vs $43.3$), $4.8$ at 9B ($49.3$ vs $44.5$), and $5.3$ at 27B.

\subsection{Certificate Mechanism Ablation}

Equal-size acceptance-rule ablations at 9B isolate what re-execution contributes (\autoref{tab:intro_rules} for the headline lift, \autoref{tab:certificate_ablation} for the per-dimension decomposition). NoCertificate is the cleanest test: it keeps VLM critique and skill routing but loosens the acceptance gate. The result is sharper than a missed lift: NoCertificate sits at Raw-level HTMLBench, only $0.4$ points higher, and its TC pass actually drops $3.1$ points below Raw ($31.0$ vs $34.1$). Removing the world-model oracle does not merely fail to help; it injects noisy supervision that hurts the very dimension downstream behavior cares about. The partial substitutes (VLM-only, Score gate, NoPreserve) recover $1.5$ to $3.6$ HTMLBench points by accepting more candidates with plausible visual changes, but each still trails the full gate by $6.6$ to $8.7$ TC-pass points. Only the full \method gate reaches $49.3$ score, $40.6$ TC, $26.1$ Func., and $66.3$ MiniApp, beating Raw-9B by $4.8$ HTMLBench and $6.5$ TC-pass points. Because all rows share the same model, recipe, and budget, the gap is evidence for the certificate rule itself.

\begin{table}[t]
\centering
\caption{SFT lift over Raw-9B per acceptance rule.}
\label{tab:intro_rules}
\setlength{\tabcolsep}{8pt}
\resizebox{\columnwidth}{!}{%
\begin{tabular}{lc}
\toprule
\textbf{Acceptance rule} & \textbf{$\Delta$ HTMLBench vs Raw-9B} \\
\midrule
\rowcolor{tablegray}
No certificate  & $+0.4$ \\
No preserve     & $+1.5$ \\
\rowcolor{tablegray}
VLM-only        & $+3.5$ \\
Score gate      & $+3.6$ \\
\rowcolor{tablegray}
\method (full)  & \textbf{+4.8} \\
\bottomrule
\end{tabular}}
\end{table}

\begin{table}[t]
\centering
\caption{Per-dimension certificate-rule ablation.}
\label{tab:certificate_ablation}
\setlength{\tabcolsep}{4pt}
\resizebox{\columnwidth}{!}{%
\begin{tabular}{lccccc}
\toprule
\textbf{Rule} & \textbf{Score} & \textbf{TC} & \textbf{Func.} & \textbf{Inter.} & \textbf{MiniApp} \\
\midrule
\rowcolor{tablegray}
VLM-only        & 48.0 & 33.4 & 23.1 & 0.4 & 62.7 \\
Score gate      & 48.1 & 34.0 & 23.4 & 0.5 & 61.4 \\
\rowcolor{tablegray}
No preserve     & 46.0 & 31.9 & 22.0 & 0.4 & 59.7 \\
No certificate  & 44.9 & 31.0 & 21.3 & 0.4 & 55.9 \\
\rowcolor{tablegray}
\method        & \textbf{49.3} & \textbf{40.6} & \textbf{26.1} & \textbf{0.6} & \textbf{66.3} \\
\bottomrule
\end{tabular}}
\vspace{-6pt}
\end{table}

\section{Analysis}
\label{sec:analysis}

\subsection{Depth and Trajectory Diagnostics}
\label{sec:analysis_depth_diagnostics}

The quality ratchet (\S\ref{sec:ratchet}) makes a structural claim about depth: each certified step strictly enlarges $\mathcal{P}_t$, so trajectories that survive deeper are not merely longer but quantitatively higher-density supervision. \autoref{tab:depth_ablation} tests that claim directly. We fix the training budget at $5{,}000$ \method examples and vary the minimum certified depth; if the ratchet works, deeper subsets should carry more verified capability per example, and if it does not, depth and signal density should decouple.

HTMLBench score rises monotonically through depth $\geq 5$ and plateaus at $\geq 8$. The headline observation is the comparison against full training: $5{,}000$ depth-$\geq 5$ examples come within $0.5$ point of the full $32{,}800$-example \method-9B on HTMLBench. Each depth-filtered example therefore carries proportionally more verified capability than an average certified-pool sample, which is what the monotone-memory property predicts: a depth-$\geq 5$ trajectory has cleared the certificate gate in sequence at least five times, each step adding to $\mathcal{P}_t$ for every later round.

\begin{table}[t]
\centering
\small
\caption{Equal-size depth diagnostic.}
\label{tab:depth_ablation}
\setlength{\tabcolsep}{6pt}
\begin{tabular}{cccccc}
\toprule
\textbf{Depth} & \textbf{Ex.} & \textbf{Score} & \textbf{TC} & \textbf{Func.} & \textbf{MiniApp} \\
\midrule
\rowcolor{tablegray}
$\geq 1$ & 5K & 43.4 & 27.3 & 19.8 & 57.8 \\
$\geq 2$ & 5K & 45.2 & 30.1 & 21.7 & 57.8 \\
\rowcolor{tablegray}
$\geq 3$ & 5K & 46.8 & 32.8 & 23.4 & 61.1 \\
$\geq 5$ & 5K & \textbf{48.8} & \textbf{36.1} & \textbf{25.5} & 58.2 \\
\rowcolor{tablegray}
$\geq 8$ & 5K & 48.5 & 35.5 & 25.1 & \textbf{61.5} \\
\bottomrule
\end{tabular}
\end{table}

Depth here counts certified rounds, not apparent runtime activity. On a held-out snapshot the runtime starts a multi-step trajectory on roughly a third of pages, but per-page certified depth collapses by more than half once the certificate gate is in force, with zero hard-verified accepted regressions. The rule axis in \S\ref{sec:experiments} and the depth axis examine the same property from two sides: the rule axis varies which gate admits, the depth axis fixes the gate and varies what we keep. Both implicate the certificate.

\subsection{Repair Needs a Verified Loop}
\label{sec:analysis_loop}

\begin{figure}[t]
\centering
\includegraphics[width=\columnwidth]{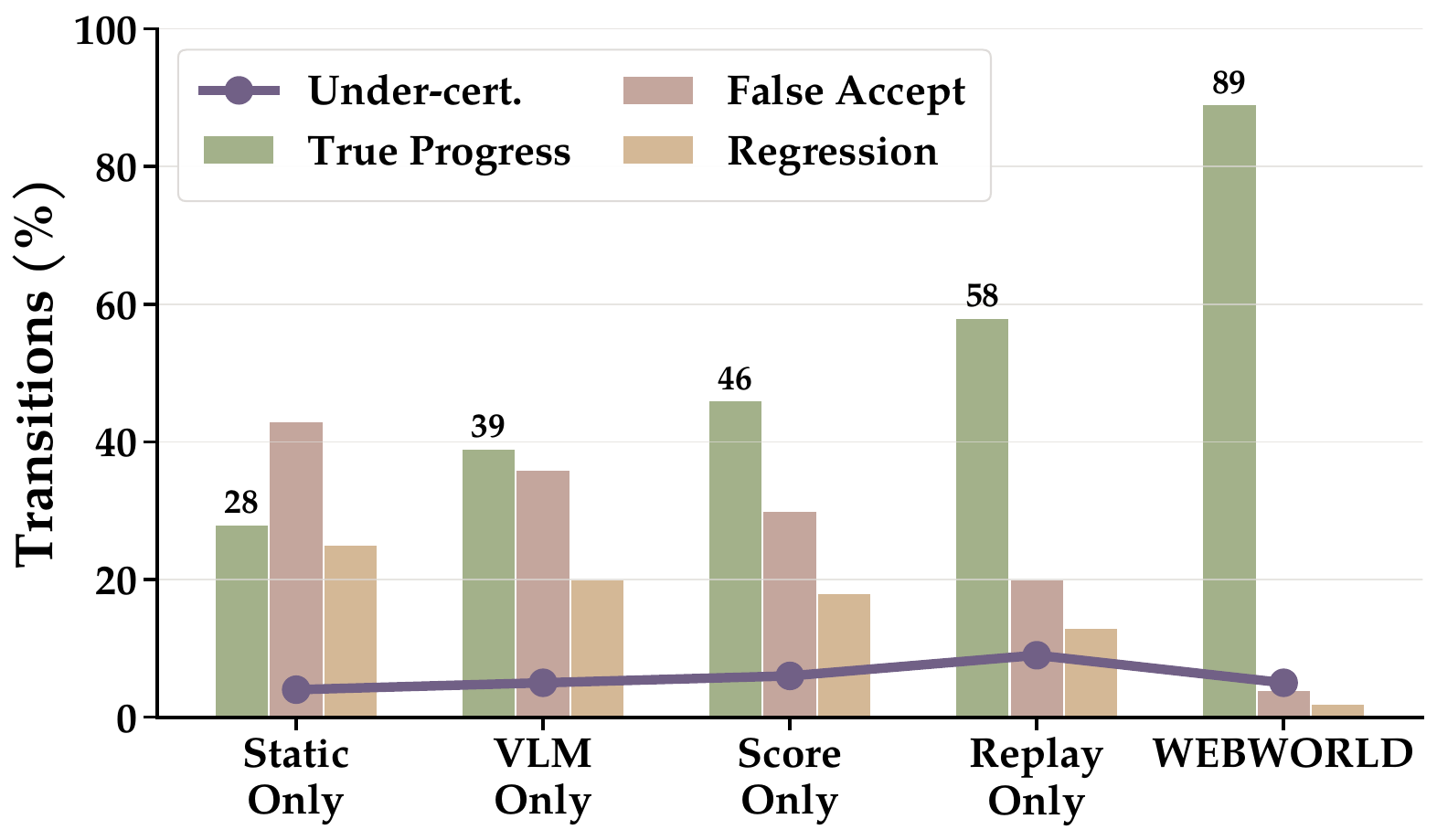}
\caption{Full admission reduces weak or false progress.}
\label{fig:analysis_admission_rules}
\vspace{-10pt}
\end{figure}

Hypothesis-proof separation (\S\ref{sec:method_overview}) demands two things from the runtime: that the proposer get repeated chances to refine its hypothesis across rounds, and that the prover keep filtering at every round. \autoref{fig:analysis_iteration_rounds} measures the first, and \autoref{fig:analysis_admission_rules} measures the second.

\autoref{fig:analysis_iteration_rounds} traces the accepted-repair distribution by round. Shallow defects clear early, harder GUI failures arrive in rounds 4--6, and the tail fades rather than growing. A one-shot loop would never see the late accepts; an unbounded one would churn past zero marginal yield. The bounded multi-round window is what lets harder failures enter the trajectory without diluting the export pool.

\autoref{fig:analysis_admission_rules} makes the dual point on the admission side: each partial substitute for the certificate captures part of the acceptance signal but leaves a residual failure mode. The four failure modes match the four lift gaps in \autoref{tab:intro_rules}. Removing the gate entirely leaves only $0.4$ points; removing only preservation leaves $1.5$; the hypothesis-only rules sit near $3.5$; only the full gate reaches $4.8$. The VLM proposes a hypothesis; replay and preservation evidence decide whether it becomes supervision.

\subsection{Verification Must Preserve Prior Capabilities}
\label{sec:analysis_preservation}

\begin{figure}[t]
\centering
\includegraphics[width=\columnwidth]{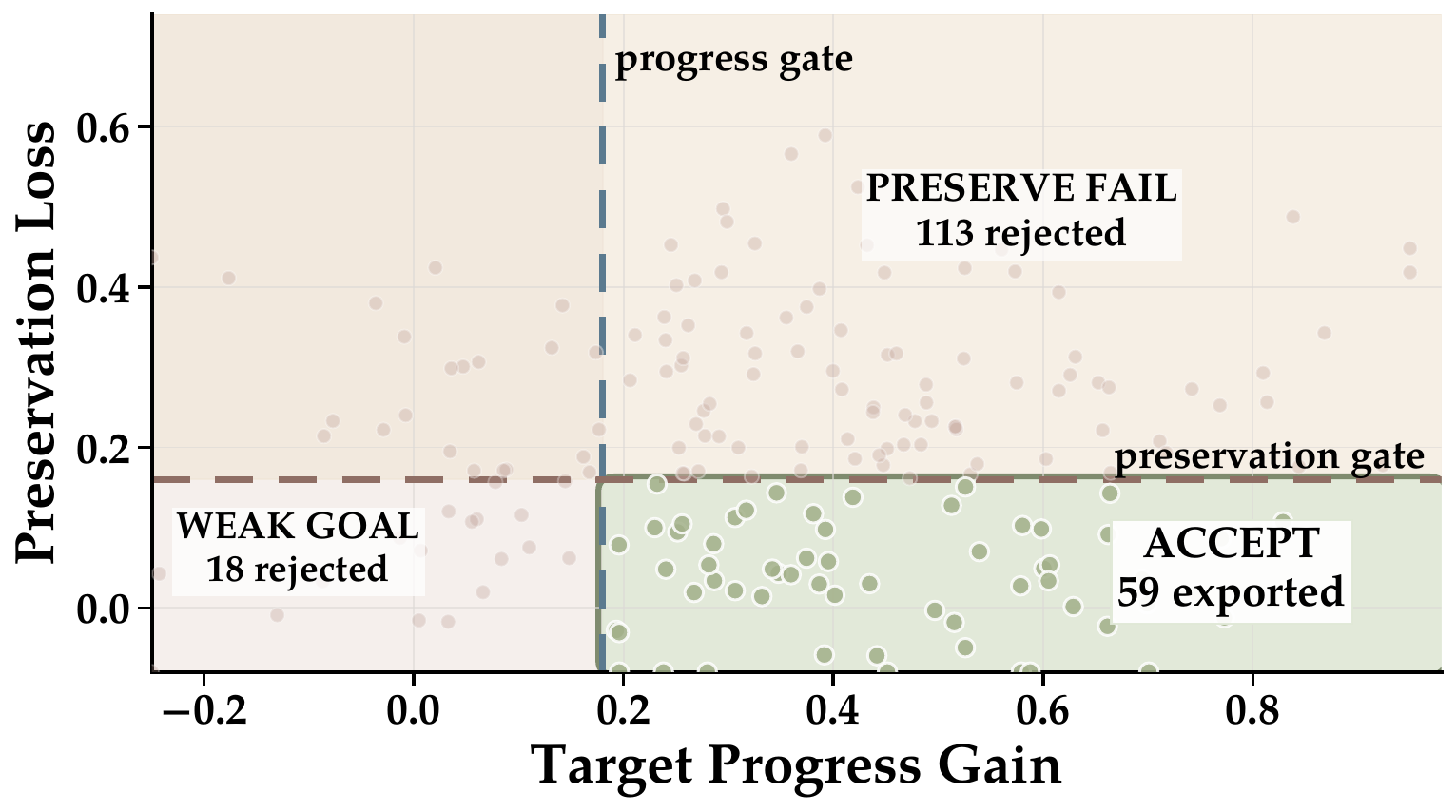}
\caption{Accepted candidates must pass both progress and preservation gates.}
\label{fig:analysis_preservation_gate}
\vspace{-8pt}
\end{figure}

\begin{figure}[t]
\centering
\includegraphics[width=\columnwidth]{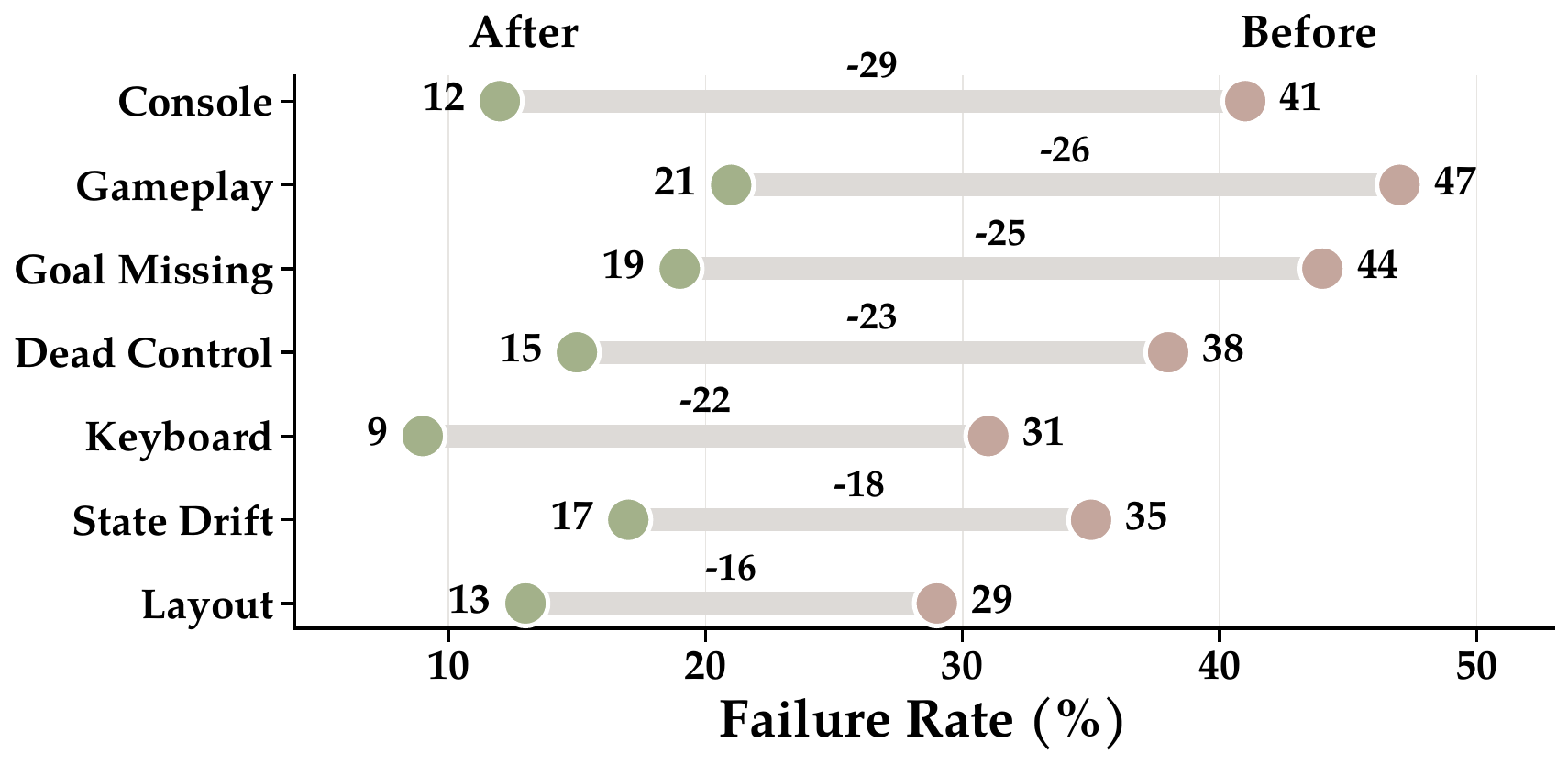}
\caption{Verified repair reduces concrete GUI failures.}
\label{fig:analysis_issue_repair_effect}
\vspace{-10pt}
\end{figure}

The third property the method claims is monotone capability memory (\S\ref{sec:method_overview}): no verified behavior may be broken later. \autoref{fig:analysis_preservation_gate} isolates the gate that enforces this property; \autoref{fig:analysis_issue_repair_effect} shows what gets through it.

A repair is admitted only if it lands in the high-progress, low-preservation-loss region; otherwise it is recorded as weak-goal or preserve-fail rejection. Web-code regressions are behavioral, not syntactic: a patch may fix layout while breaking keyboard control, or polish a canvas while erasing a working goal condition. Type checks, unit tests, and screenshot scores miss these cases; inherited preserve-set replay is required.

\autoref{fig:analysis_issue_repair_effect} traces the downstream consequence: console errors, gameplay failures, dead controls, keyboard failures, and layout breaks all decrease after verification-backed repair, while preserve regressions stay near zero. The cost of removing this gate is visible in \S\ref{sec:experiments}: NoPreserve still accepts progress but loses $3.3$ HTMLBench points to the full gate, with the gap concentrated on TC pass and Functionality where regressions hide. The loop becomes a ratchet rather than a cosmetic filter.

\section{Related Work}
\label{sec:related}

\textbf{Web agents and front-end evaluation.}
Language-agent work interleaves reasoning, actions, and feedback in external environments: ReAct introduces reasoning--action traces, WebShop grounds agents in a shopping environment, and Reflexion adds verbal feedback across trials~\citep{yao2023react,yao2022webshop,shinn2023reflexion}. Browser benchmarks extend this to realistic websites and multimodal browsing, including Mind2Web, WebArena, WebVoyager, and VisualWebBench~\citep{deng2023mind2web,zhou2024webarena,he2024webvoyager,liu2024visualwebbench}. Front-end generation and evaluation target code rather than an existing interface: Design2Code and WebSight study screenshot-to-code~\citep{si2024design2code,laurencon2024websight}, Image2Struct studies structure extraction~\citep{roberts2024image2struct}, DesignBench, FullFront, and WebGen-Bench test broader interactive generation~\citep{xiao2025designbench,sun2025fullfront,lu2025webgenbench}, and ArtifactsBench evaluates rendered multimodal artifacts~\citep{zhang2025artifactsbench}. These lines use the browser as an environment to navigate or as an oracle for a render score. Our method instead treats the browser as a world model for code: each repair asks it to issue or refuse a transition certificate, so supervision is certified rather than a rollout or screenshot.

\textbf{Repair, judging, and self-improvement.}
Feedback-driven repair revises generations from execution or critique signals. Self-Refine, Self-Debugging, CodeRL, and self-repair studies all turn execution or critique into a revision loop~\citep{madaan2023selfrefine,chen2023selfdebugging,le2022coderl,olausson2024selfrepair}, while SWE-bench frames real-world repair as resolving issue reports against a repository with tests~\citep{jimenez2024swebench}. SWE-bench works because the test suite is an oracle the patch cannot rewrite; most web-code defects, however, are visual or interactional and lack an equivalent suite. Synthetic-data and unsupervised code-generation work explores alternatives to labeled corpora~\citep{gunasekar2023textbooks,wu-etal-2026-ucoder}, and exploration-driven web pipelines such as Explorer synthesize or filter agent trajectories~\citep{pahuja2025explorer}. MLLM-as-judge frameworks~\citep{zhang2025artifactsbench} try to play that oracle role with a visual judge, but the judge inherits the proposer's modality bias, the failure mode our intro audit measures. It confines the VLM to hypothesis generation and gives the oracle role to the browser: re-execution under a typed interaction contract must certify target progress and preservation. Separated hypothesis and proof make accepted transitions SFT data, not another repair step.

\paragraph{Concurrent related work.}
HTMLCure is concurrent work on tests and state-aware repair for interactive HTML, and provides the HTMLBench-400 testbed used here~\citep{htmlcure2026}. \method uses that testbed only for held-out evaluation. Its contribution is different: browser re-execution becomes the world-model counterparty, critiques become interaction contracts, and only certified transitions are exported for SFT.

\section{Conclusion}
\label{sec:conclusion}

The structural failure of VLM-driven web-code self-improvement is that the proposer is also the judge, so the loop optimizes for visually plausible outputs that often fail under interaction. The method removes that conflation by routing every accepted transition through a counterparty the VLM cannot fool: the browser, treated as a deterministic and executable world model for web code. Hypothesis and proof are split by construction. The VLM proposes via a typed interaction contract; only re-execution under that contract can issue the acceptance certificate that lets a candidate become the next baseline. Certified transitions then accumulate into the agent's verified-capability memory and the SFT export.

Under matched backbone and training recipe, the resulting certified trajectories train a 27B model that improves Raw-27B by \textbf{5.3} points on HTMLBench-400 and \textbf{14.9} points on MiniAppBench-Val, reaching Kimi-K2.6/GPT-5.4-level performance in this interactive HTML setting. Equal-size acceptance-rule ablations show the gain is carried by world-model-backed admission itself: dropping the certificate collapses the lift to $0.4$ points over Raw-9B, more than ten times below the full gate.

The central design choice is the interface, not a stronger prompt or critic. The browser replays intended interactions, checks target predicates, and preserves earlier capabilities; only those certified transitions update memory and enter SFT. Training with the full gate improves behavior, while removing the certificate removes the gain. Together, method, data construction, and ablation point to the same mechanism: browser admission as a world model makes supervision reliable.

\clearpage
\section*{Limitations}
\method currently isolates single-file interactive HTML artifacts, where browser behavior can be controlled and the certificate effect measured directly. Broader multi-file or framework-heavy settings add deployment variables such as dependency replay, project organization, and agent scaffolding. The hypothesis-proof separation also depends on the browser being a faithful oracle for the property at stake. Genuinely perceptual defects, such as aesthetic taste, brand identity, or accessibility nuance, have no executable certificate and fall back to the VLM proxy, inheriting its noise. The verifier reduces false positives but does not make VLM critique infallible: a wrong critique can still waste repair budget, and dynamic applications still need deterministic replay.

\section*{Ethics Statement}
The system edits executable web code and therefore inherits standard risks of code generation, including accidental insecure scripts, misleading interfaces, or broken accessibility behavior. We restrict the current setting to sandboxed HTML artifacts and use verification gates to reject regressions before exporting data. Before release, generated HTML artifacts and trajectories are scanned for private credentials, personal identifiers, external network endpoints, malicious scripts, and offensive content; flagged examples are removed. Any deployment of repaired artifacts should still include security, privacy, and accessibility review.

\section*{Acknowledgments}
This work was supported by the National Natural Science Foundation of China (62525601), CCF-Huawei Populus Grove Fund, the Fundamental Research Funds for the Central Universities (Grant No.~GW2025-19), and the State Key Laboratory of Complex \& Critical Software Environment (Grant No.~SKLCCSE-2025ZX-26).

\bibliography{custom}
\bibliographystyle{acl_natbib}

\clearpage
\appendix
\section{HTMLBench Benchmark Details}
\label{app:benchmark}

We evaluate \method on two external interactive-HTML benchmarks. \textbf{HTMLBench-400}~\citep{htmlcure2026} is a frozen single-file benchmark with deterministic browser test cases. \textbf{MiniAppBench-Val}~\citep{zhang2026miniappbench} is a held-out mini-application set used as an out-of-distribution transfer check. Each HTMLBench item is a natural-language prompt scored along five dimensions: rendering, visual design, functionality, interactivity, and code quality. The scores sum to $100$. Functionality is the dominant component and is measured as the weighted pass rate over a frozen pool of \textbf{6{,}000} deterministic browser test cases. \autoref{tab:bench_families} gives the six task families and their item and test-case counts. The benchmark has 65 subtypes spread across 122 easy, 156 medium, and 122 hard prompts; 338 prompts require interaction.

\begin{table*}[t]
\centering
\small
\setlength{\tabcolsep}{6pt}
\caption{HTMLBench-400 task families. Each family contains natural-language prompts paired with deterministic browser test cases used to score functionality.}
\label{tab:bench_families}
\begin{tabular}{lcccccc|c}
\toprule
& \textbf{Apps \& Tools} & \textbf{Content \& Mkt.} & \textbf{Data Viz.} & \textbf{Games \& Sim.} & \textbf{3D / WebGL} & \textbf{Visual Art} & \textbf{Total} \\
\midrule
Items        & 105 & 110 & 35 & 55 & 20 & 75 & \textbf{400} \\
Test cases   & 1{,}688 & 1{,}660 & 588 & 682 & 328 & 1{,}054 & \textbf{6{,}000} \\
Subtypes     & 18 & 16 & 7 & 9 & 4 & 11 & \textbf{65} \\
\bottomrule
\end{tabular}
\end{table*}

Test cases are deterministic browser programs built from actions such as click, type, hover, key press, resize, visibility check, JavaScript assertion, and screenshot-change check. They are prompt-grounded and avoid framework-specific selectors, hidden source assumptions, real credentials, payments, and private services. A page's functionality score is the weighted pass rate over the test pool; coverage is recorded only as execution metadata. The visual-design component is the only one assigned through a VLM path (over curated post-execution keyframes); the remaining four dimensions are static or browser-execution checks. We therefore report total score and TC pass side by side so that visual-only gains are not misread as functional progress.

\subsection{Leakage Audit}
\label{app:leakage_audit}

Because HTMLBench-400 and \method share infrastructure ancestry through HTMLCure, we verify five non-overlap conditions explicitly:

\noindent\textbf{(i) Item disjointness.}
The source HTML artifacts that seed the roughly $60$K-candidate trajectory loop are drawn from a separate pool of front-end generation requests. No HTMLBench-400 item ID, prompt string, rendered DOM, or screenshot enters the data-construction pipeline. The two sets are checked by exact prompt-string hash and by item-identifier comparison; the intersection is empty.

\noindent\textbf{(ii) Prompt-template disjointness.}
The VLM critique prompt operates over the runtime's observation $o_t$ (screenshots, DOM facts, console, probe outcomes); it is not seeded with any HTMLBench-400 prompt template, rubric, or scoring instruction. The two prompt families share no text.

\noindent\textbf{(iii) Test-case disjointness.}
HTMLBench-400's $6{,}000$ deterministic test cases are authored against benchmark items by a separate team and are not accessible to the training loop. \method's contract grammar generates per-page predicates from the VLM critique; these predicates are not derived from, and do not consult, any HTMLBench test case.

\noindent\textbf{(iv) Probe set disjointness.}
\method's $14$ probe macros are generic interaction patterns (click, drag, keyboard, form fill, hover, etc.); they are not specialized to HTMLBench item structure and can fire against any rendered page. HTMLBench-400 test cases are item-specific scripted assertions. The two libraries share no procedure or selector vocabulary.

\noindent\textbf{(v) Repair-skill disjointness.}
The fifteen VLM repair skills and five deterministic patch lanes route by issue family (gameplay, control wiring, SVG, structural, visual, maintenance); none is tied to an HTMLBench task family, item ID, or test case. Removing or extending the skill set changes which contracts get patched, not which contracts get certified.

The HTMLCure analysis runner is shared infrastructure used only at evaluation time, where it scores held-out HTMLBench-400 items against trained checkpoints; it is never queried during data construction or training. Under these five conditions, any gain we report on HTMLBench-400 reflects transfer from certified supervision, not benchmark leakage.

\section{Artifacts, Licenses, and Intended Use}
\label{app:artifacts}

\method uses or creates four artifact classes: the browser-verification runtime (contracts, probes, certificates, and typed rejection logs), the certified transition corpus, SFT checkpoints, and external evaluation artifacts. HTMLBench-400 and the HTMLCure runner are cited as the held-out evaluation benchmark and runner; MiniAppBench-Val is cited as the out-of-distribution transfer benchmark. The web-agent and web-code benchmarks discussed in \S\ref{sec:related} are cited for comparison only and are not used to construct the training corpus.

External benchmarks, model checkpoints, and model APIs are used for research evaluation under their documented access conditions. The \method artifacts are intended for research on browser-verified web-code generation and should not be used to deploy generated web code without independent security, privacy, and accessibility review. We plan to release code, logs, and data only in forms compatible with the underlying artifact terms; generated trajectories are filtered as described in the Ethics Statement before release.

\section{SFT Training Details}
\label{app:sft_plan}

All SFT routes use the same conversation template, maximum sequence length, optimizer recipe, and evaluation runner within a model scale. The main comparison trains Raw and \method variants at 4B, 9B, and 27B. Mechanism runs are performed on the 9B backbone with equal-size route construction so that the acceptance rule is the changing factor. Depth runs also keep the example count fixed, which separates trajectory quality from sample quantity. This controlled-route design makes the causal comparison the construction of supervision rather than model family, prompt format, or evaluation protocol.

Backbone is the Qwen3.5 family at 4B/9B/27B parameters. Optimizer is AdamW with learning rate $2{\times}10^{-5}$, weight decay $0.01$, and a $3\%$ linear warmup. Effective batch is $128$ via gradient accumulation; sequence length is $16{,}384$ tokens; each route is trained for three epochs. Hardware is $64{\times}$H100 for 4B and 9B routes and $128{\times}$H100 for 27B routes. We report compute in H20-equivalent card-hours, using the accounting conversion $1$ H100 card-hour $=4$ H20 card-hours. One 4B route uses $64{\times}4$ H100 card-hours, or $1{,}024$ H20-equivalent card-hours; one 9B route uses $64{\times}6$ H100 card-hours, or $1{,}536$ H20-equivalent card-hours; one 27B route uses $128{\times}12$ H100 card-hours, or $6{,}144$ H20-equivalent card-hours. All cells in \autoref{tab:main_sft_results}, \autoref{tab:intro_rules}, \autoref{tab:certificate_ablation}, and \autoref{tab:depth_ablation} are single-seed runs with seed $42$.

Evaluation is the public HTMLCure analysis runner pinned to its release commit; the MiniAppBench-Val evaluation uses the released validation split with the runner's default settings. Browser execution uses a headless Chromium/Playwright runtime with deterministic viewport and timeout settings. VLM inference for critique uses greedy decoding (temperature $0$, max output $4{,}096$ tokens); patch generation uses temperature $0.7$ with one candidate per contract. Training uses the same software environment across routes; exact package versions, runner commit, and checkpoint identifiers are released with the code artifact.

\section{Data Construction Pipeline}
\label{app:funnel_details}

This appendix gives the engineering details abstracted in \autoref{fig:funnel}. The runtime processes each source artifact through up to $T_{\max}=10$ critique-repair-verify rounds. Within a round, the VLM emits one structured critique; the planner attempts to admit it as a contract; if admitted, the routed skill produces one candidate patch; the runtime re-executes the candidate under the contract. The round ends in one of two states: a certificate is issued and the candidate becomes the next baseline, or a typed rejection feeds the next critique.

Across the runtime we observed, roughly $60$K admitted repair candidates reached the browser re-execution stage after contract filtering. Re-execution refused $24{,}350$ candidates with typed rejection reasons such as partial progress, visual-only evidence, preserve risk, or stale replay. The certified pool that resulted is $42{,}860$ transitions, decomposed by proof level into same-trace replay ($42.3\%$), capability gain ($22.5\%$), target-issue progress ($16.4\%$), localized visual evidence ($11.5\%$), and static structural repair ($7.3\%$). The final $32{,}800$-example SFT corpus is a uniform random subsample (seed $42$) of this certified pool, matching the Raw budget.

\section{Supplementary Analysis}
\label{app:supp_analysis}

The main analysis focuses on the three mechanism-level claims: repair should be iterative, admission should be verification-backed, and GUI evidence should cover interactive behavior rather than static appearance alone. This appendix reports lower-level diagnostics from the same evaluation. These figures are not separate benchmark claims; they explain where the loop fails, why \method rejects many visually plausible candidates, and which evidence paths make an accepted transition auditable.

\subsection{Failure Modes and Rejection Accounting}
\label{app:failure_accounting}

\begin{figure}[t]
\centering
\includegraphics[width=\columnwidth]{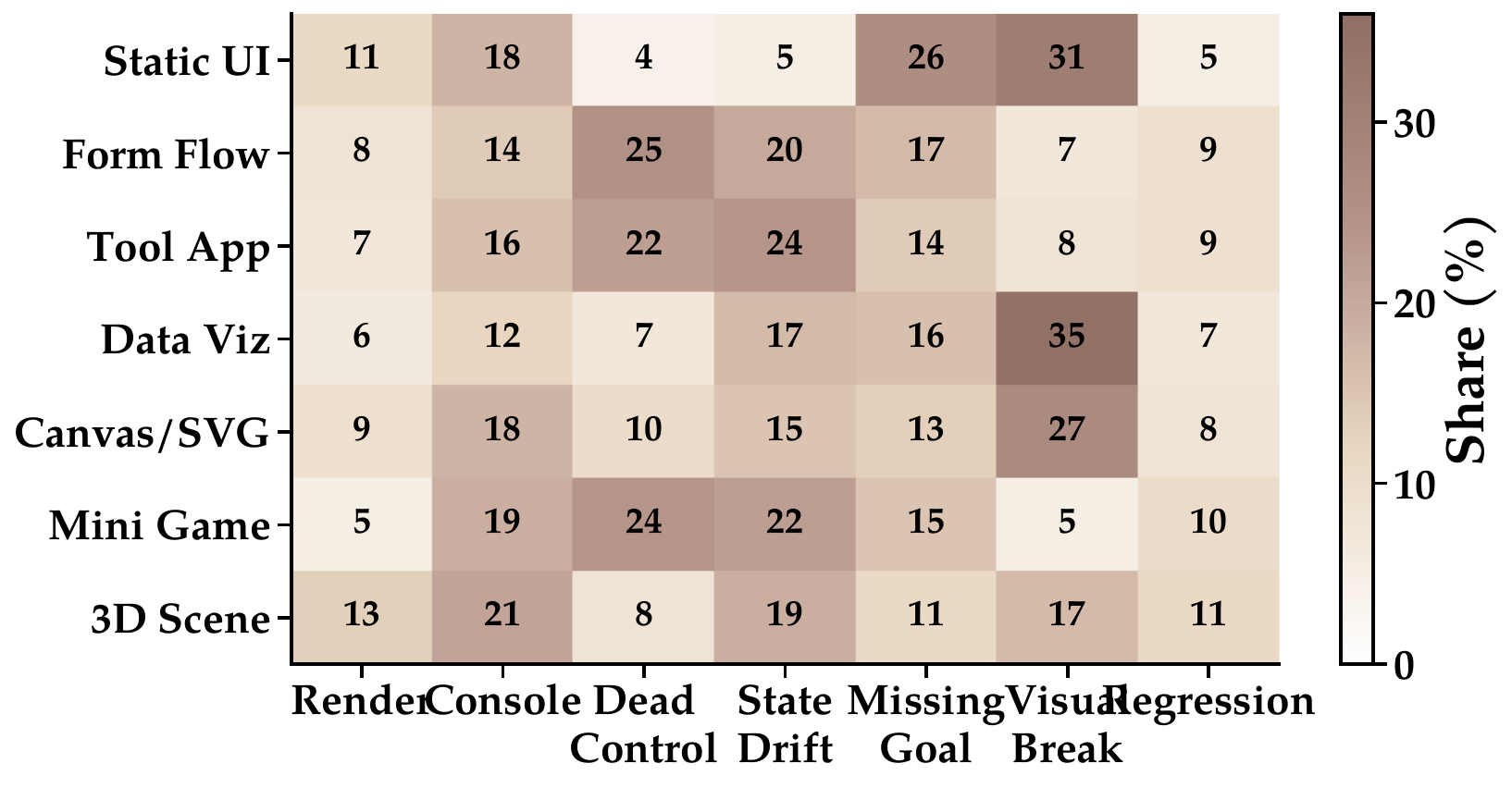}
\caption{Failure modes vary by page type.}
\label{fig:app_failure_taxonomy}
\end{figure}

\autoref{fig:app_failure_taxonomy} motivates using executed GUI traces instead of a single aggregate score. A static UI, a form workflow, a tool application, a canvas/SVG page, a mini game, and a 3D scene do not fail in the same way. Some failures are visible at render time, but others require clicks, typed input, keyboard control, state replay, or console inspection. This supports the design choice in \method: the VLM critique is grounded in the observed failure family, and the repair contract narrows the next edit to the missing behavior rather than to generic visual polish.

\begin{figure}[t]
\centering
\includegraphics[width=\columnwidth]{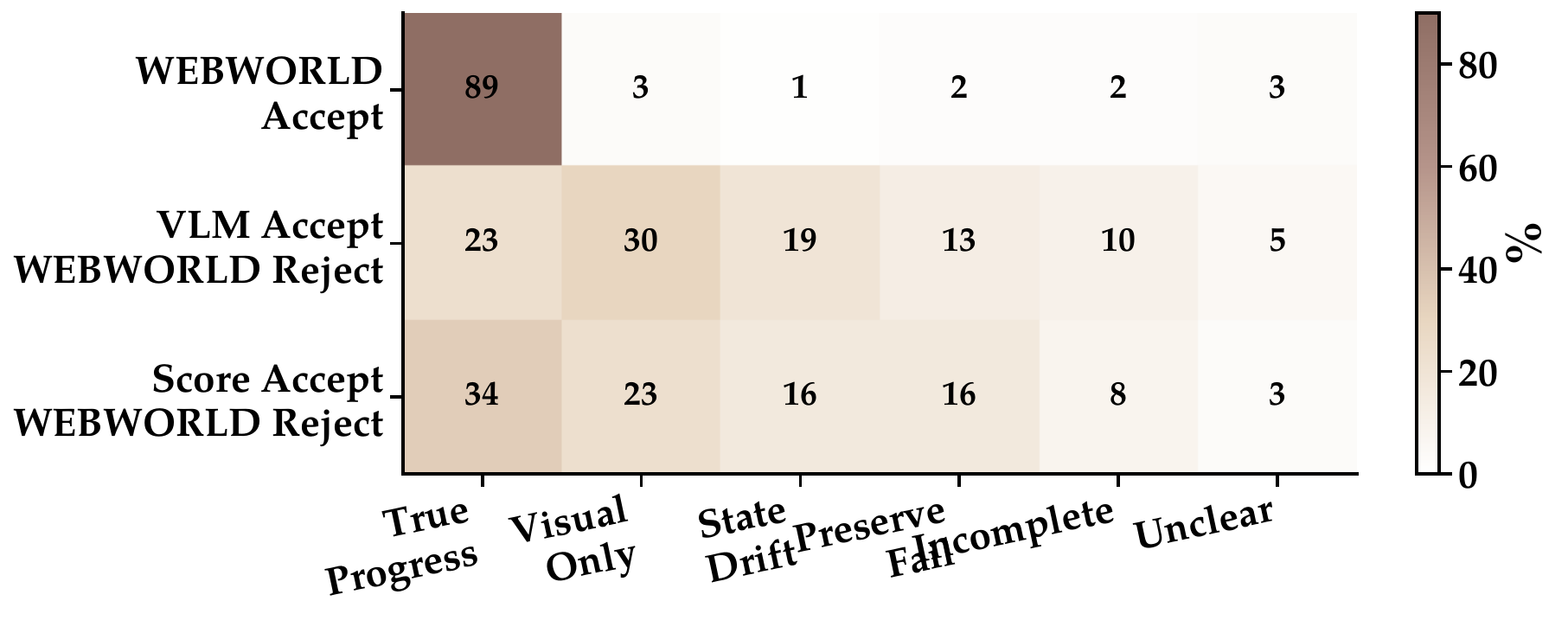}
\caption{Full admission rejects common false positives.}
\label{fig:app_accept_reject_audit}
\end{figure}

\autoref{fig:app_accept_reject_audit} explains why the paper treats VLM judgment as a hypothesis rather than proof. VLM-only and score-only rules still accept candidates that look improved but fail under replay, drift because the state changed independently, or lack enough preservation evidence. The full \method gate rejects those cases and exports mostly true-progress transitions. This is the practical reason for the quality-ratchet contract: rejected attempts remain useful diagnostics, but they should not become SFT targets.

\subsection{Evidence Paths and Repair Accounting}
\label{app:evidence_paths}

\begin{figure}[t]
\centering
\includegraphics[width=\columnwidth]{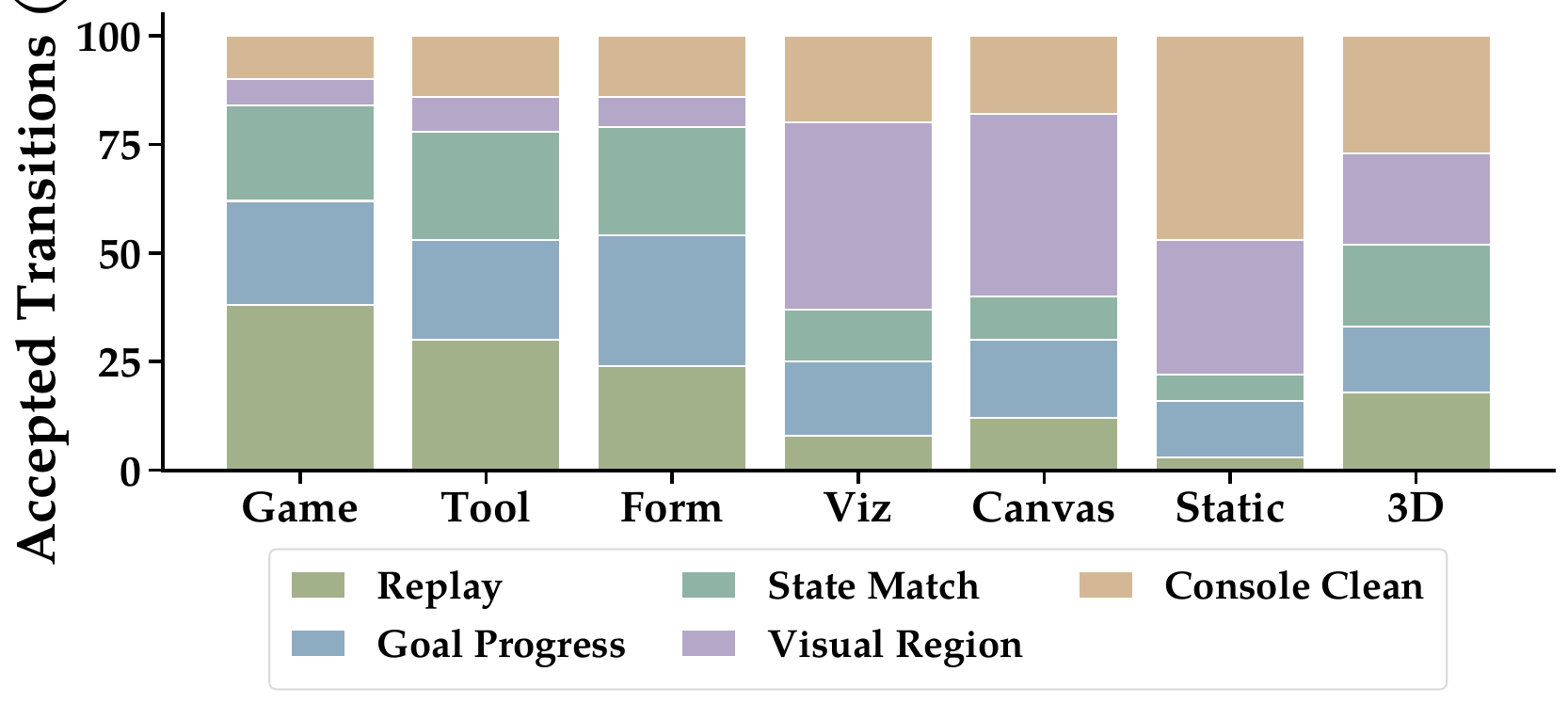}
\caption{Accepted transitions use diverse evidence paths.}
\label{fig:app_evidence_paths}
\end{figure}

\autoref{fig:app_evidence_paths} shows that accepted transitions are not admitted through one brittle test. Game pages rely heavily on replay and goal progress; forms and tools need state matching and workflow evidence; visual and static pages use visual-region evidence only when it is tied to a localized target and low preserve risk. This diversity matters because web-code quality is multimodal. A valid repair may fix control wiring, state machines, visual regions, or console/runtime cleanup. The common requirement is not the evidence type, but that the evidence proves target progress under the repair contract.

\begin{figure}[t]
\centering
\includegraphics[width=\columnwidth]{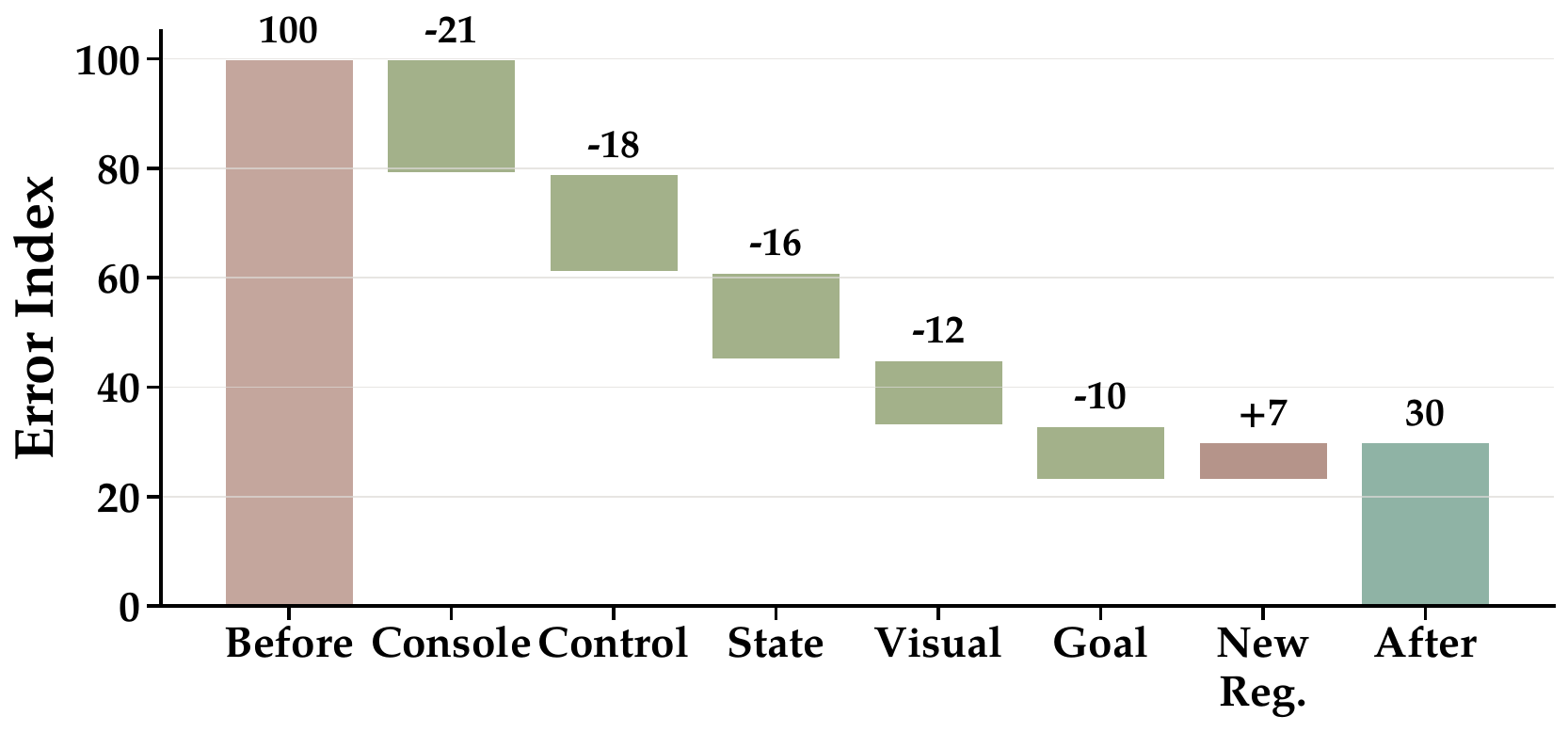}
\caption{Verified repair reduces the aggregate error index.}
\label{fig:app_repair_waterfall}
\end{figure}

\autoref{fig:app_repair_waterfall} summarizes the same process as an error-index accounting view. Most of the reduction comes from fixing concrete GUI failures: console/runtime defects, dead controls, state mismatches, visual breaks, and missing goals. The small new-regression bar is the residual risk the preservation gate is designed to expose. The point is therefore calibrated: \method does not claim that every edit is safe. It claims that re-execution and preservation-aware admission make the remaining exported transitions much cleaner than a raw critique-and-rewrite loop.

\subsection{Cost, Frontier, and Case Trajectories}
\label{app:cost_cases}

\begin{figure}[t]
\centering
\includegraphics[width=\columnwidth]{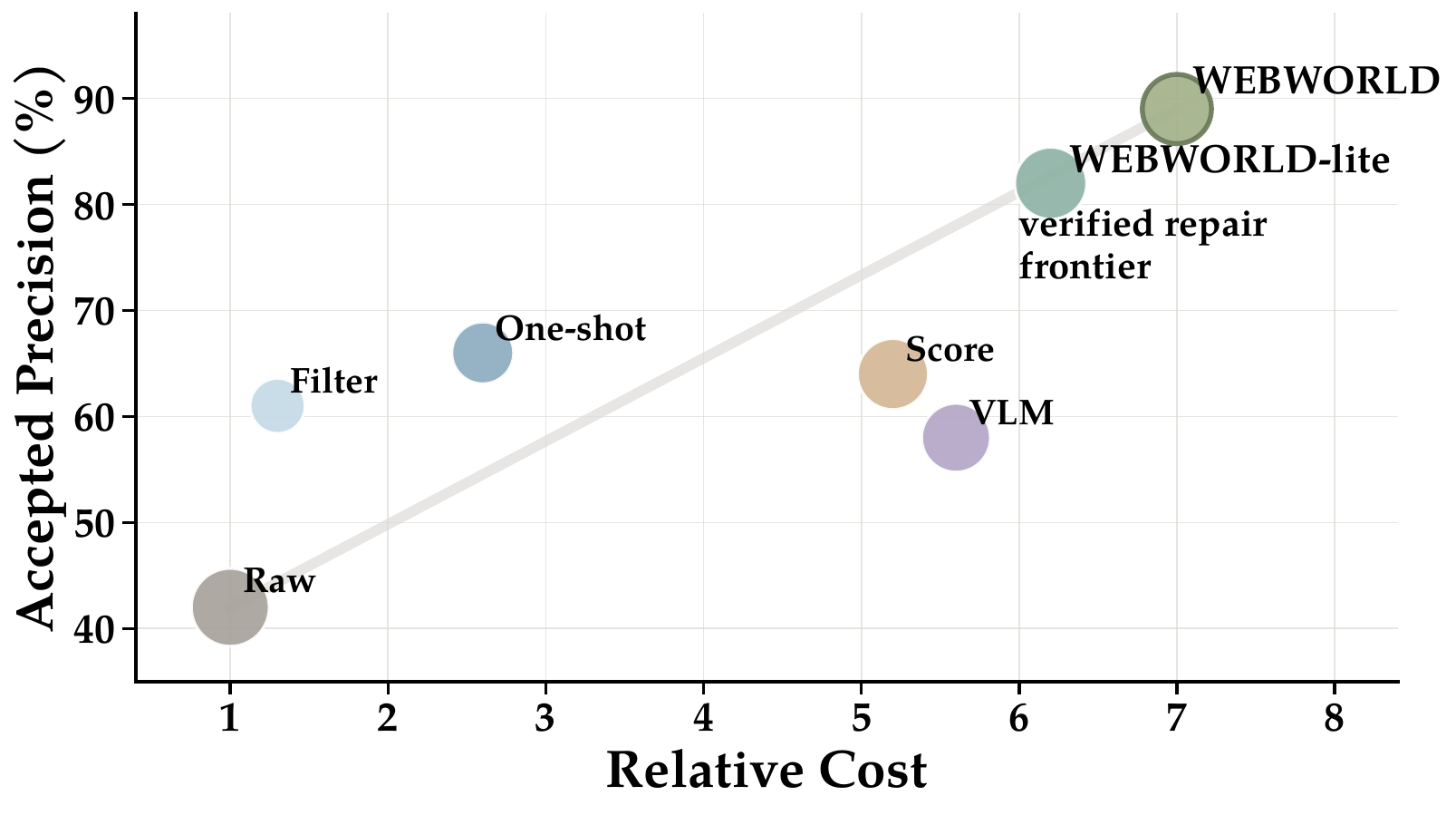}
\caption{Verified repair trades additional cost for cleaner accepts.}
\label{fig:app_cost_quality}
\end{figure}

\autoref{fig:app_cost_quality} places the mechanism in a cost-quality space. Cheap filtering and one-shot repair are useful baselines, but they do not provide enough accepted precision for clean trajectory export. VLM-only and score-only routes spend more than filtering while still underperforming because their acceptance signal is not tied tightly enough to replay and preservation. \method moves to the verified-repair frontier. It costs more because it runs interaction, critique, patching, and verification, but that cost buys cleaner accepted transitions. This is the intended tradeoff for SFT data construction, where noisy positives are costlier than refusals.

\begin{figure}[t]
\centering
\includegraphics[width=\columnwidth]{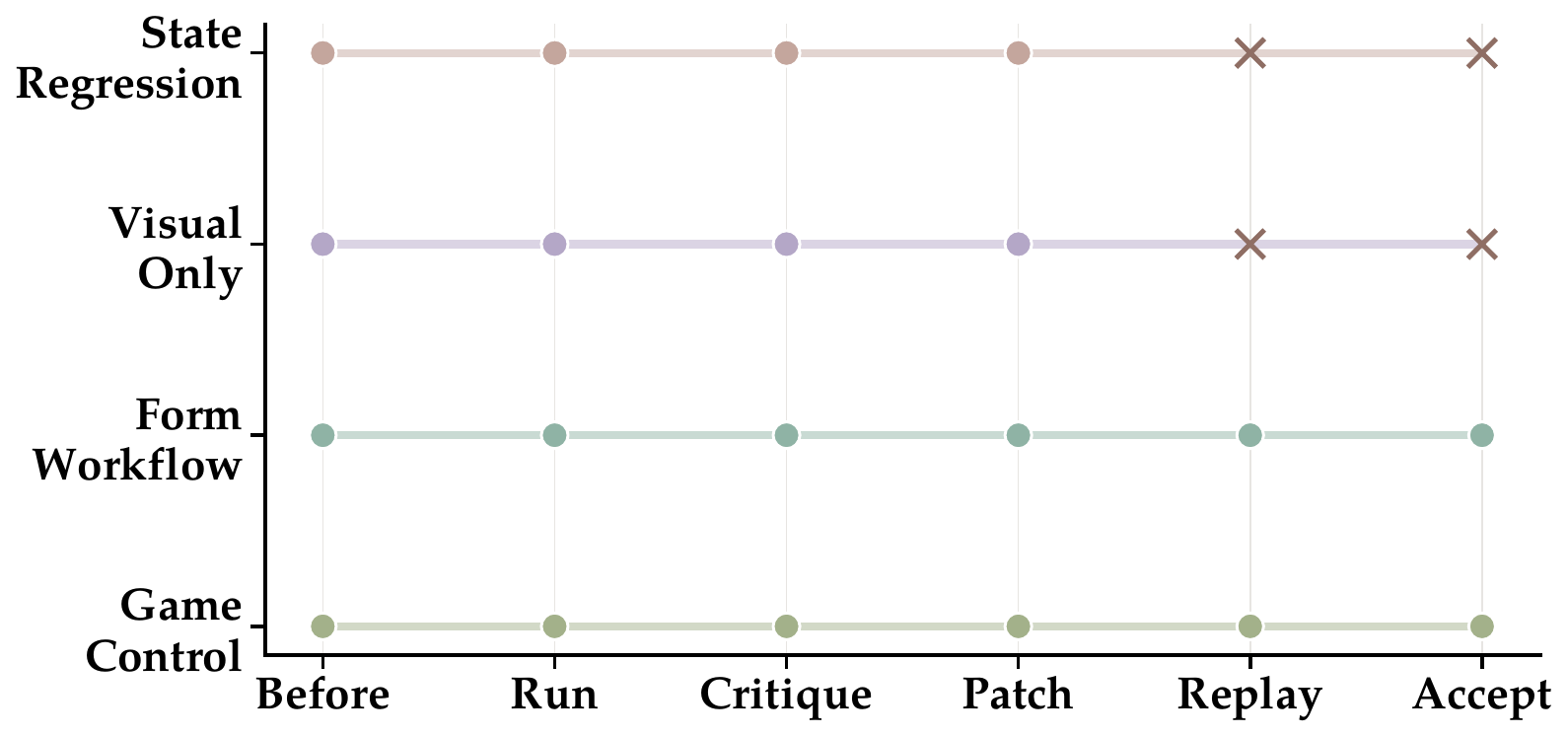}
\caption{Case timelines show accepted and rejected repair paths through the verifier.}
\label{fig:app_case_timeline}
\end{figure}

\autoref{fig:app_case_timeline} gives the qualitative counterpart to the aggregate diagnostics. Accepted cases move through browser execution, critique, patch generation, replay, and final acceptance. Rejected cases can survive early stages but fail when replay exposes state regression or when the evidence remains visual-only. This is the behavior desired from an autonomous repair loop: the system may explore many candidate edits, but only transitions with replayable progress and preserved prior capability enter the verified trajectory.

\section{Rendered Case Studies}
\label{app:rendered_cases}

This section expands the trajectory view into rendered browser states. The appendix remains in the normal two-column format: each screenshot is shown as one rendered state, followed immediately by the corresponding interpretation. Each figure makes the same repair contract visible: browser replay must establish progress, the gate accepts or rejects the candidate, and preservation locks keep prior evidence intact. This layout emphasizes the core contract of \method: visually plausible but non-replayable edits are rejected.

\newcommand{\caseimage}[3]{%
\par\medskip\noindent
\begin{minipage}{\columnwidth}
\centering
\includegraphics[width=\columnwidth]{#1}\par
\refstepcounter{figure}\label{#2}%
\vspace{2pt}{\small\textbf{Figure~\thefigure:} #3\par}%
\end{minipage}\par}

\newcommand{\caseannot}[2]{%
\par\nopagebreak\vspace{2pt}\noindent
{\small\sloppy\setlength{\emergencystretch}{6em}%
\textbf{Contract.}\,#1\par
\textbf{Certificate.}\,#2\par}\vspace{2pt}}

\subsection{Case A: Zoo Tycoon GUI Repair}
\label{app:case_truck_balance}

Each figure below shows only the browser-rendered product HTML. The replay verifier for the same round is generated as a separate artifact, so the visual case study stays faithful to the user-facing GUI.

\caseimage{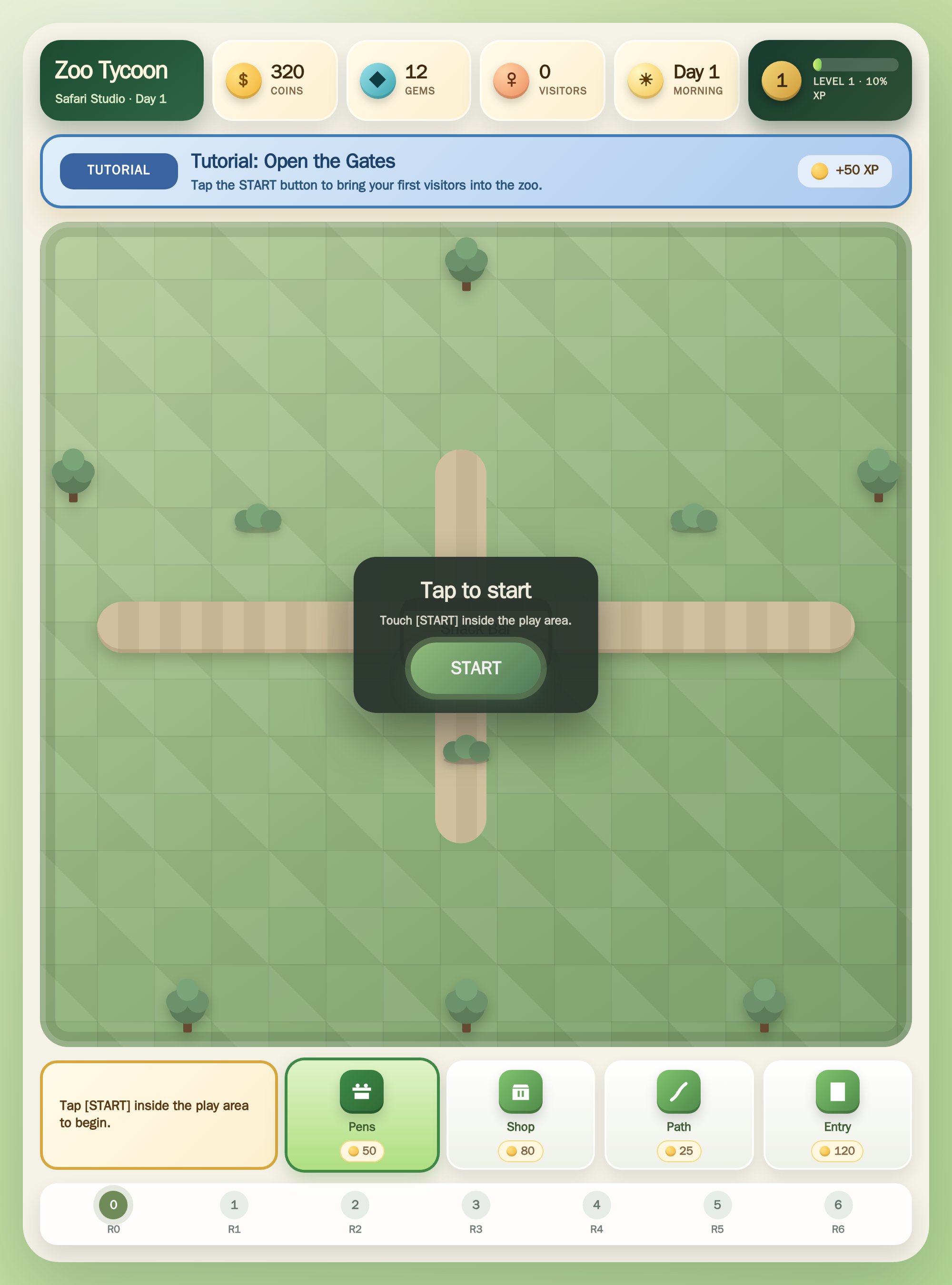}{fig:case_truck_00}{Zoo Tycoon initial state.}
\caseannot{(no contract yet --- this is the unverified baseline page)}{no certificate; the page has not been re-executed under any contract}
\noindent\textbf{Interpretation.}
The initial product page has a complete zoo scene with animal pens, visitor paths, economy cards, and a guide modal. The missing piece is evidence: the start/guide state has not yet been replayed as a concrete browser target, so the separate verifier keeps it as a baseline.

\caseimage{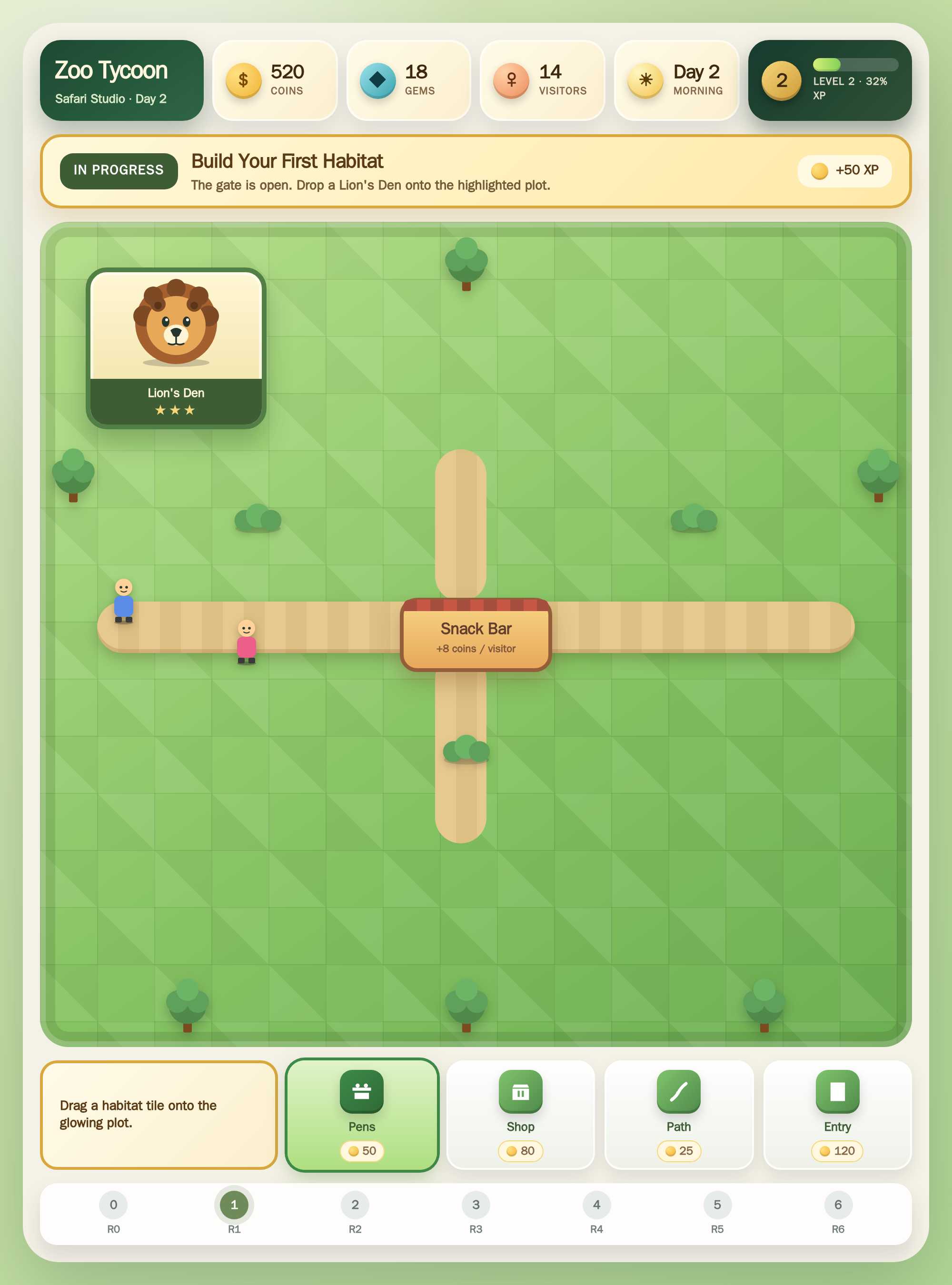}{fig:case_truck_01}{Round 1 accepted: semantic target resolves.}
\caseannot{\texttt{target}\,=\,\texttt{state\_changed}\allowbreak\texttt{(button.}\allowbreak\texttt{guide-start)};\ \texttt{replay}\,=\,\texttt{click}\allowbreak\texttt{(button.guide-start)}\allowbreak\texttt{+screenshot};\ \texttt{preserve}\,=\,\{\texttt{button.}\allowbreak\texttt{guide-start.replayed}\}}{\textsc{accepted} via same-trace replay proof}
\noindent\textbf{Interpretation.}
Round 1 keeps the zoo scene intact while turning the guide action into a visible, clickable target. This is the first synchronized repair: the product screenshot exposes the target, and the separate replay record can resolve it.

\caseimage{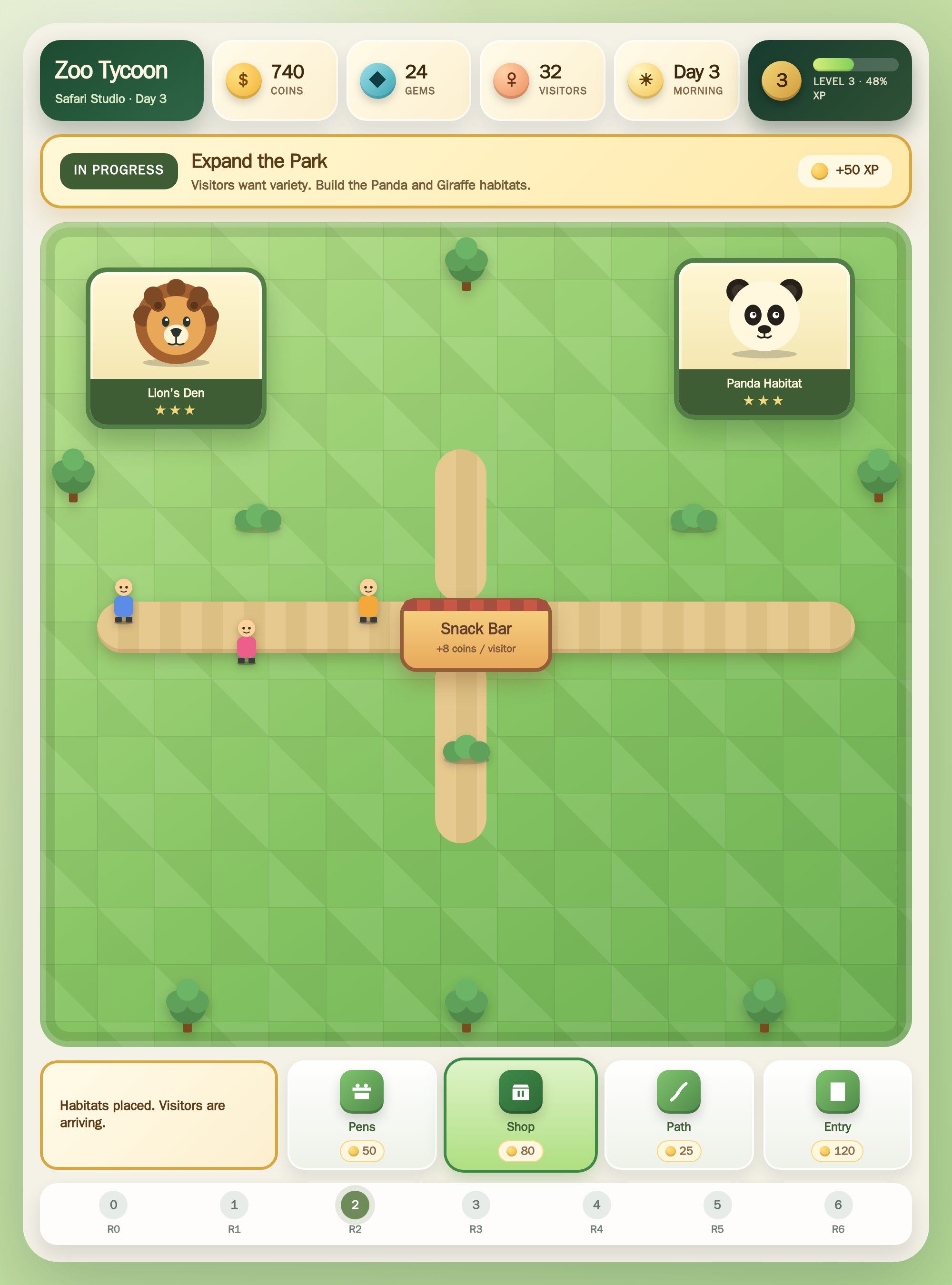}{fig:case_truck_02}{Round 2 accepted: native zoo feedback.}
\caseannot{\texttt{target}\,=\,\texttt{habitat-region.populated};\ \texttt{replay}\,=\,\texttt{rerun(layer-1-2)}\allowbreak\texttt{+inspect(zoo-grid)};\ \texttt{preserve}(2)\,=\,prev\,$\cup$\,\{\texttt{habitat-region}\}}{\textsc{accepted} via capability gain proof; a new probe macro now passes}
\noindent\textbf{Interpretation.}
Round 2 improves the HTML itself: additional animal habitats, path structure, and build tools become easier to read. The accepted repair is therefore a native GUI refinement, not hidden metadata.

\caseimage{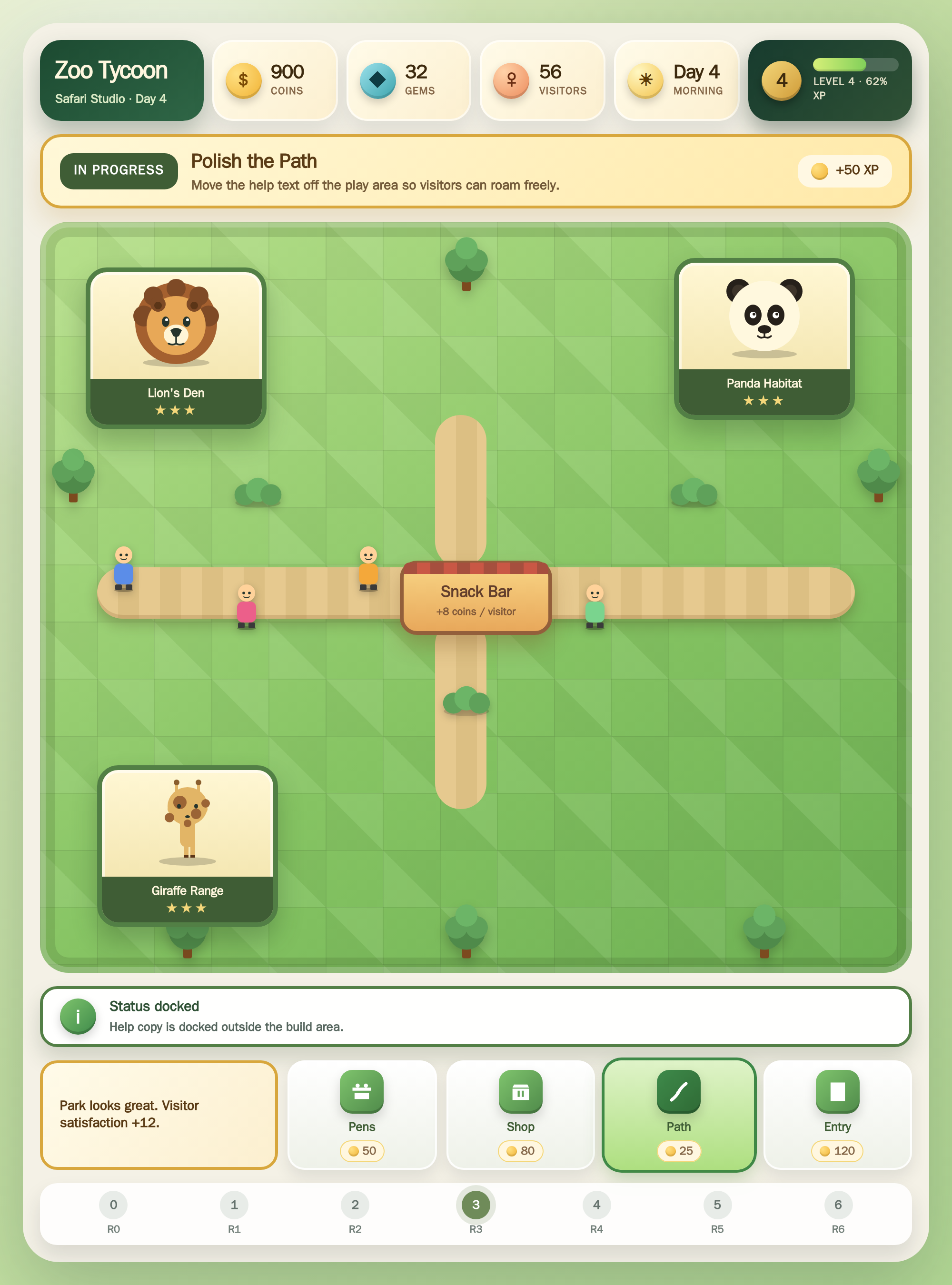}{fig:case_truck_03}{Round 3 accepted: status copy preserved.}
\caseannot{\texttt{target}\,=\,\texttt{status-copy}\,$\notin$\,\texttt{play-area};\ \texttt{replay}\,=\,\texttt{dom-snapshot+visual-cmp};\ \texttt{preserve}(3)\,=\,prev\,$\cup$\,\{\texttt{status-copy.docked}\}}{\textsc{accepted} via localized visual proof under bounded region risk}
\noindent\textbf{Interpretation.}
Round 3 docks clearer status/help copy beside the board while keeping working controls and animals visible. The improvement is visual clarity under preservation, not an overlay that hides the task.

\caseimage{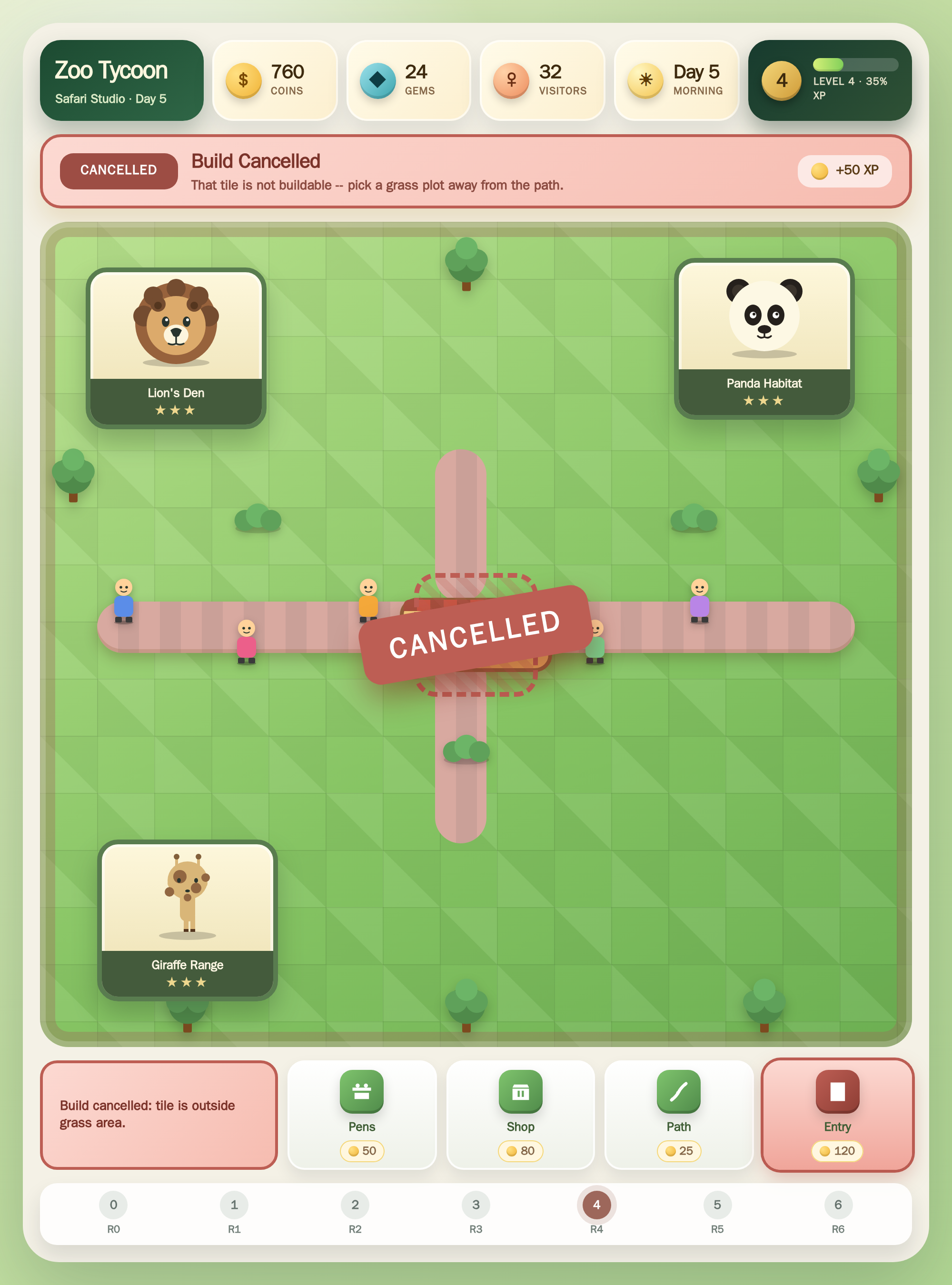}{fig:case_truck_04}{Round 4 rejected: wrong semantic target.}
\caseannot{\texttt{target}\,=\,\texttt{wrong-control}\allowbreak\texttt{.click\_feedback};\ \texttt{replay}\,=\,\texttt{click(.snack-bar)}\allowbreak\texttt{+screenshot};\ \texttt{preserve}\,=\,(unchanged from R3)}{\textsc{rejected} --- the candidate is anchored to the wrong semantic control; replay never reaches the start gate, so no certificate is issued}
\noindent\textbf{Interpretation.}
Round 4 shows the negative case. The product page remains plausible, but the candidate points to the wrong semantic control. The rejection is shown in the status and round marker rather than as a board overlay, so the verifier decision is visible without covering the zoo scene.

\caseimage{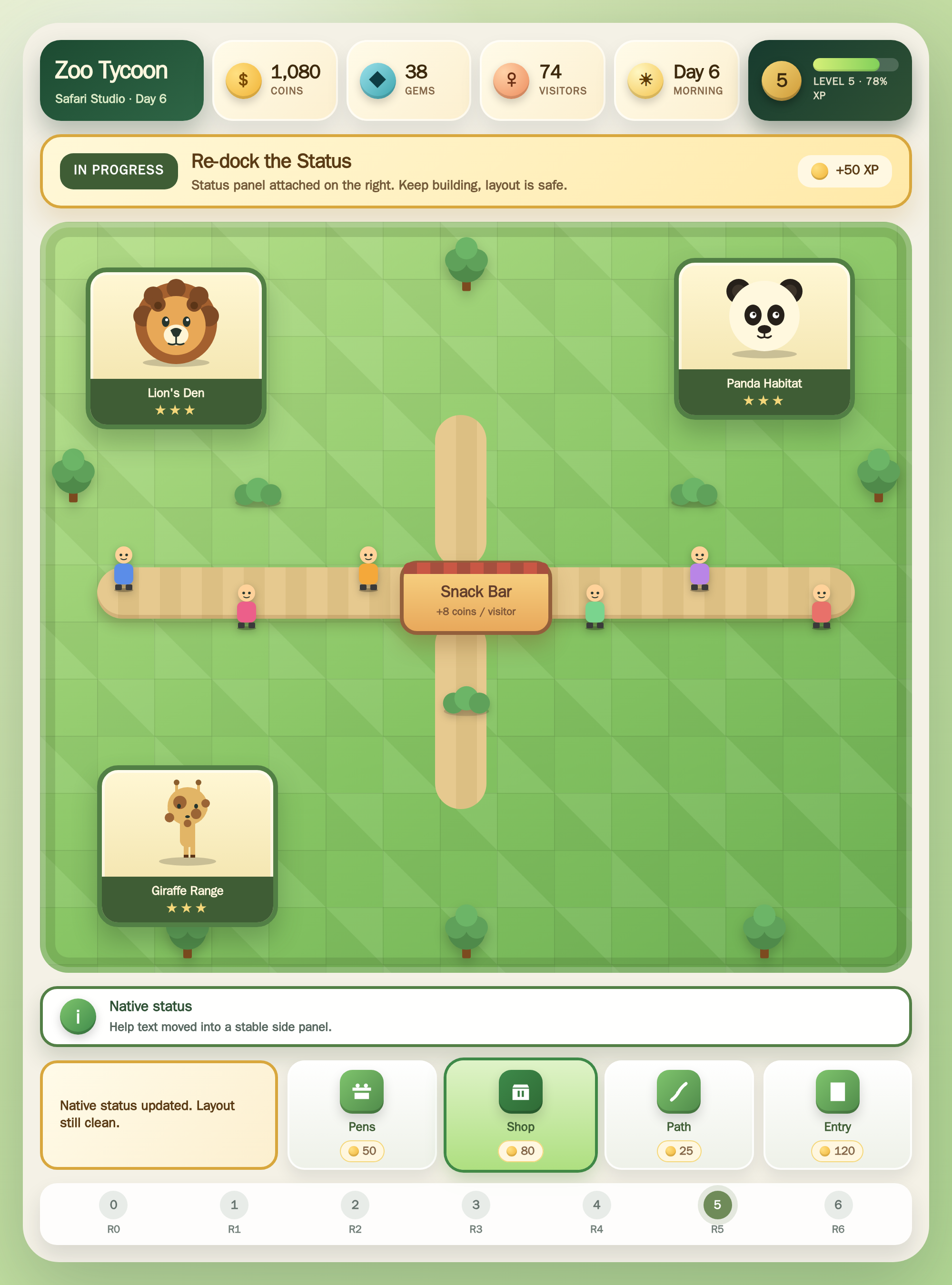}{fig:case_truck_05}{Round 5 accepted: native status block.}
\caseannot{\texttt{target}\,=\,\texttt{native-status.visible}\,$\land$\,no-overlay;\ \texttt{replay}\,=\,\texttt{rerun+visual-cmp};\ \texttt{preserve}(4)\,=\,prev\,$\cup$\,\{\texttt{native-status.visible}\}}{\textsc{accepted} via localized visual proof}
\noindent\textbf{Interpretation.}
Round 5 repairs the previous issue by moving guidance into a stable native status region. The visual change is deliberate: feedback is easier to scan, but the board and toolbar remain intact, so the preserved evidence still matches the screenshot.

\caseimage{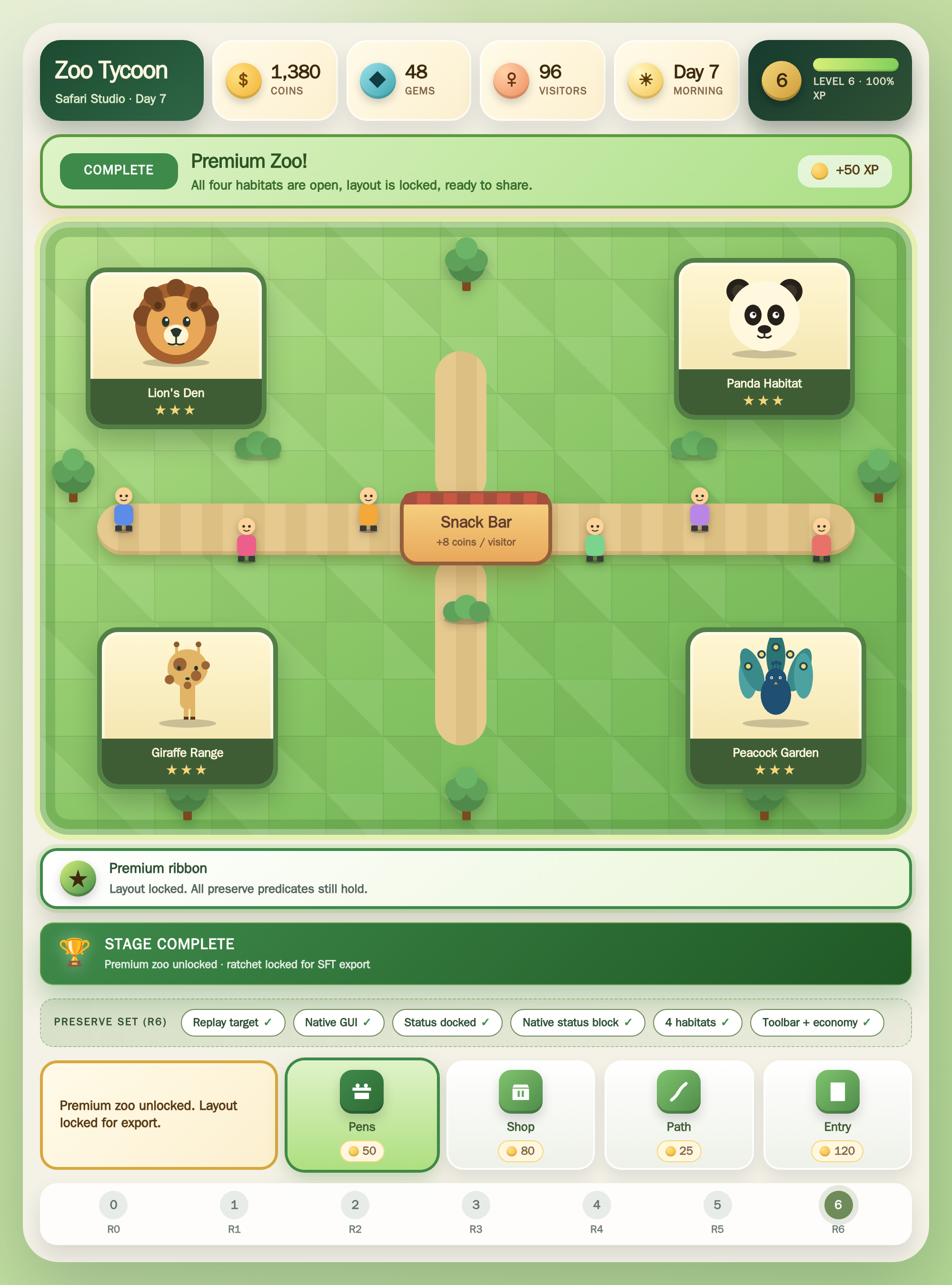}{fig:case_truck_06}{Round 6 accepted: exportable zoo state.}
\caseannot{\texttt{target}\,=\,all-predicates-hold;\ \texttt{replay}\,=\,\texttt{full-rerun};\ \texttt{preserve}(6)\,=\,prev\,$\cup$\,\{\texttt{toolbar.on},\allowbreak\,\texttt{economy-card.stable}\}}{\textsc{export-ready} --- every preserve predicate still holds and the certified trajectory enters the SFT export pool}
\noindent\textbf{Interpretation.}
Round 6 is the exportable state. The HTML is more attractive and more structured, and the separate verifier explains why it is safe to keep: the replay target, native grid, status text, and visible content are all locked.

\subsection{Case B: Smart Finance Filter Repair}
\label{app:case_finance_filter}

\caseimage{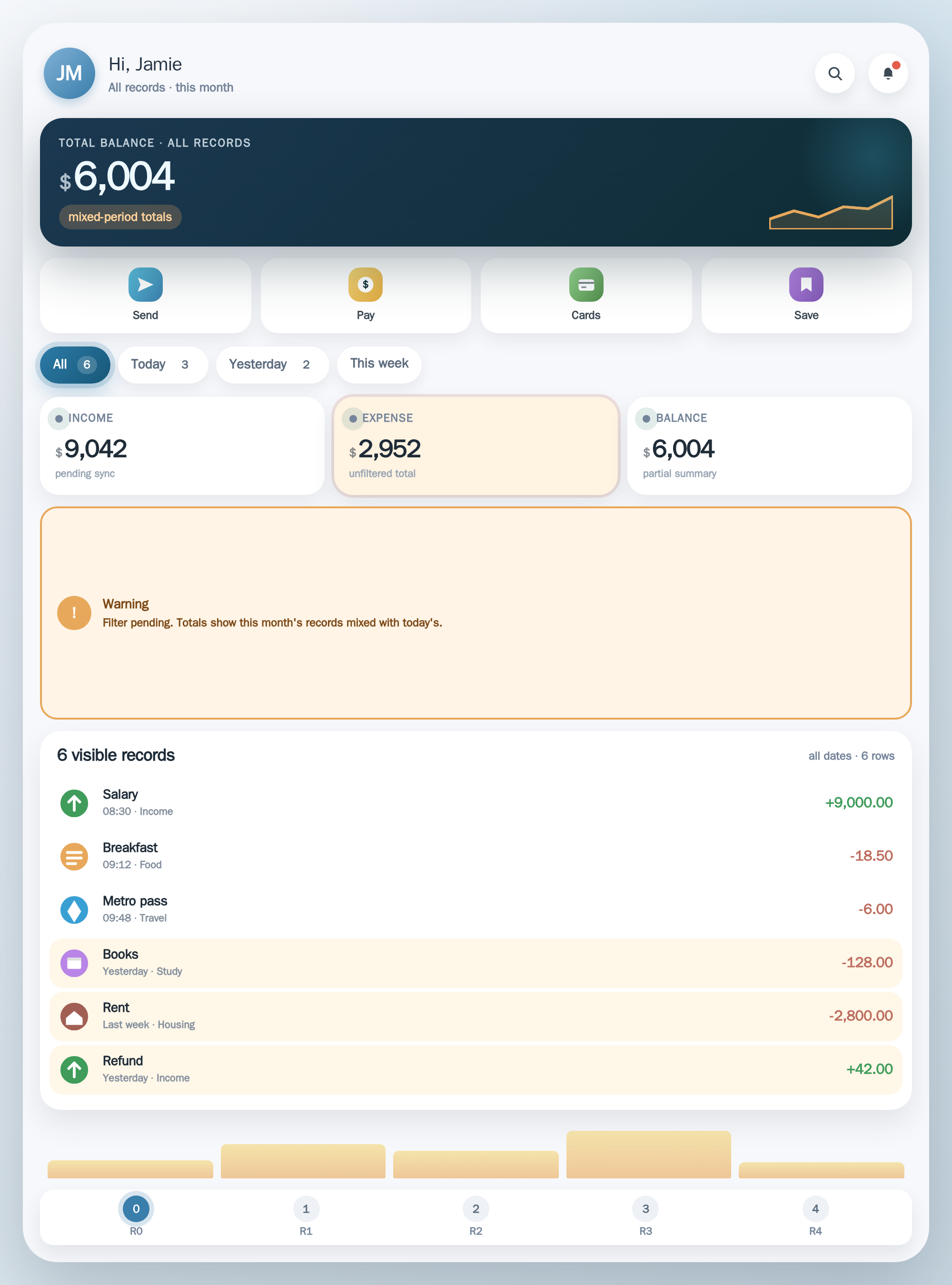}{fig:case_finance_00}{Smart Finance initial state.}
\caseannot{(no contract yet --- this is the unverified baseline page)}{no certificate; the page has not been re-executed under any contract}
\noindent\textbf{Interpretation.}
The initial finance example has a clean sidebar, filter chips, balance cards, and a transaction table. At this point the layout is present, but the filter is only visual; the verifier has not yet observed a replayed state transition.

\caseimage{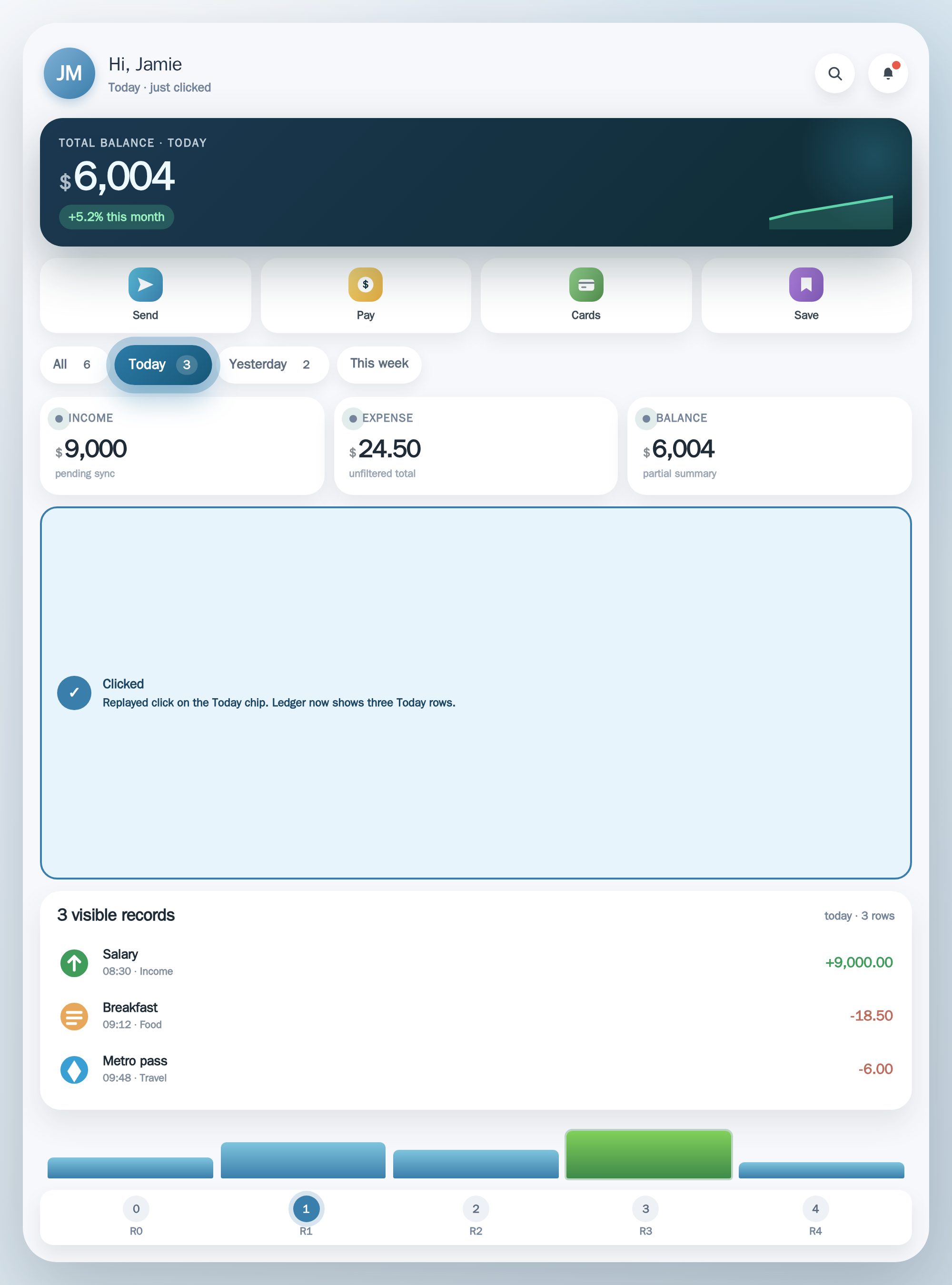}{fig:case_finance_01}{Round 1 accepted: Today-filter click.}
\caseannot{\texttt{target}\,=\,\texttt{button:`Today'.active};\ \texttt{replay}\,=\,\texttt{click(button:`Today')+screenshot};\ \texttt{preserve}(1)\,=\,\{\texttt{button:`Today'.active}\}}{\textsc{accepted} via same-trace replay proof}
\noindent\textbf{Interpretation.}
Round 1 makes the Today chip visibly active and binds it to the replay trace. The screenshot and text now describe the same change: a static filter row becomes a verified interactive state.

\caseimage{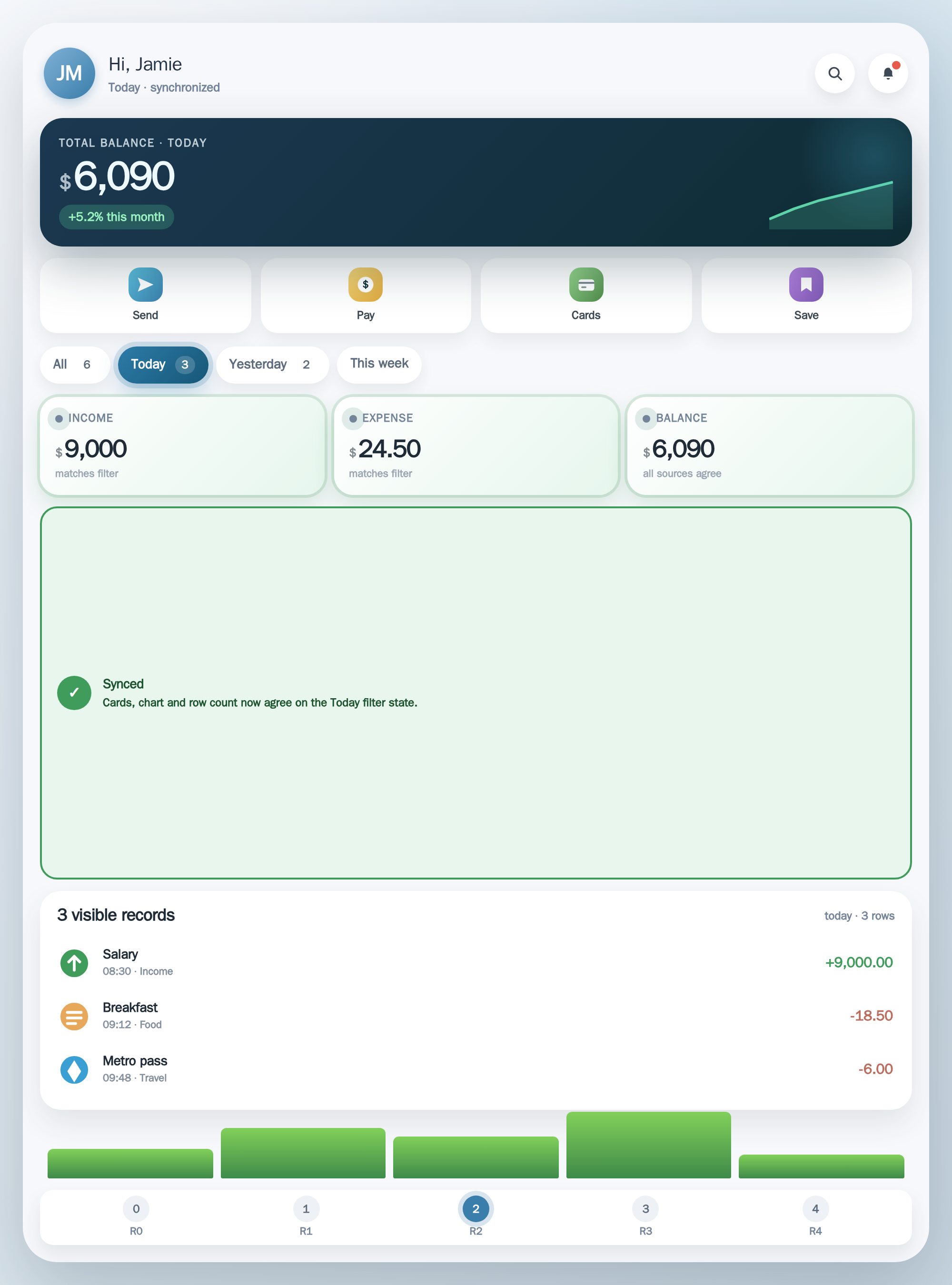}{fig:case_finance_02}{Round 2 accepted: cards synchronized.}
\caseannot{\texttt{target}\,=\,\texttt{cards.totals}\,$=$\,\texttt{filter.rows};\ \texttt{replay}\,=\,\texttt{rerun(layer-1-3)}\allowbreak\texttt{+probe(totals)};\ \texttt{preserve}(2)\,=\,prev\,$\cup$\,\{\texttt{cards.synced}\}}{\textsc{accepted} via capability gain proof; cards/rows/chart now agree on the filtered state}
\noindent\textbf{Interpretation.}
Round 2 improves both correctness and aesthetics: the cards, compact chart, and row count now share the same filtered state. This prevents a weak cosmetic repair where only the tab color changes.

\caseimage{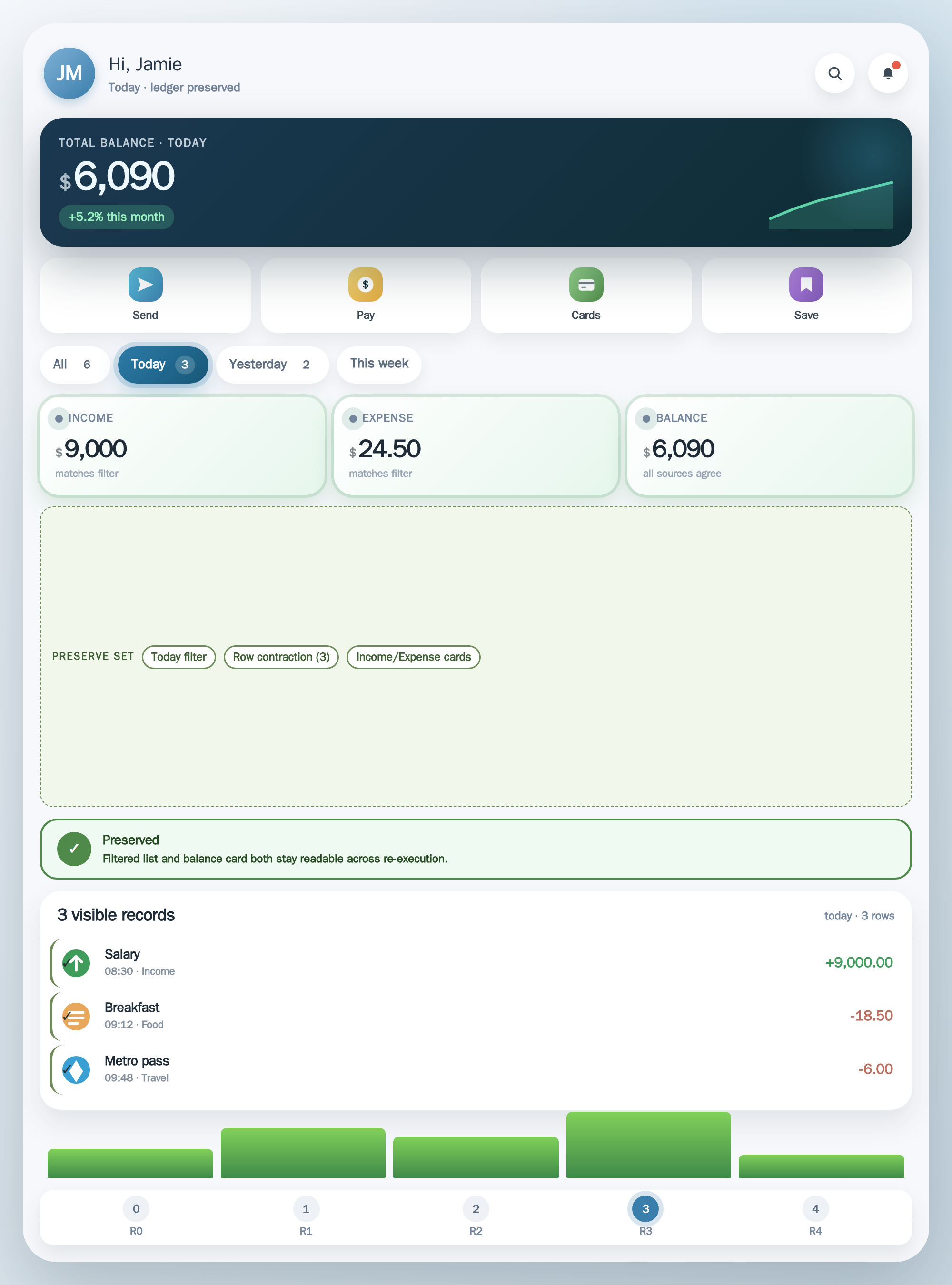}{fig:case_finance_03}{Round 3 accepted: lists preserved.}
\caseannot{\texttt{target}\,=\,\texttt{ledger.readable}\allowbreak\,$\land$\,\texttt{filter.locked};\ \texttt{replay}\,=\,\texttt{dom-snapshot}\allowbreak\texttt{+visual-cmp};\ \texttt{preserve}(3)\,=\,prev\,$\cup$\,\{\texttt{ledger.preserved}\}}{\textsc{accepted} via localized visual proof with bounded region risk contained}
\noindent\textbf{Interpretation.}
Round 3 keeps the transaction list readable while preserving the filtered dashboard. The visual improvement is the stable information hierarchy: cards summarize the state, and rows still provide the evidence behind it.

\caseimage{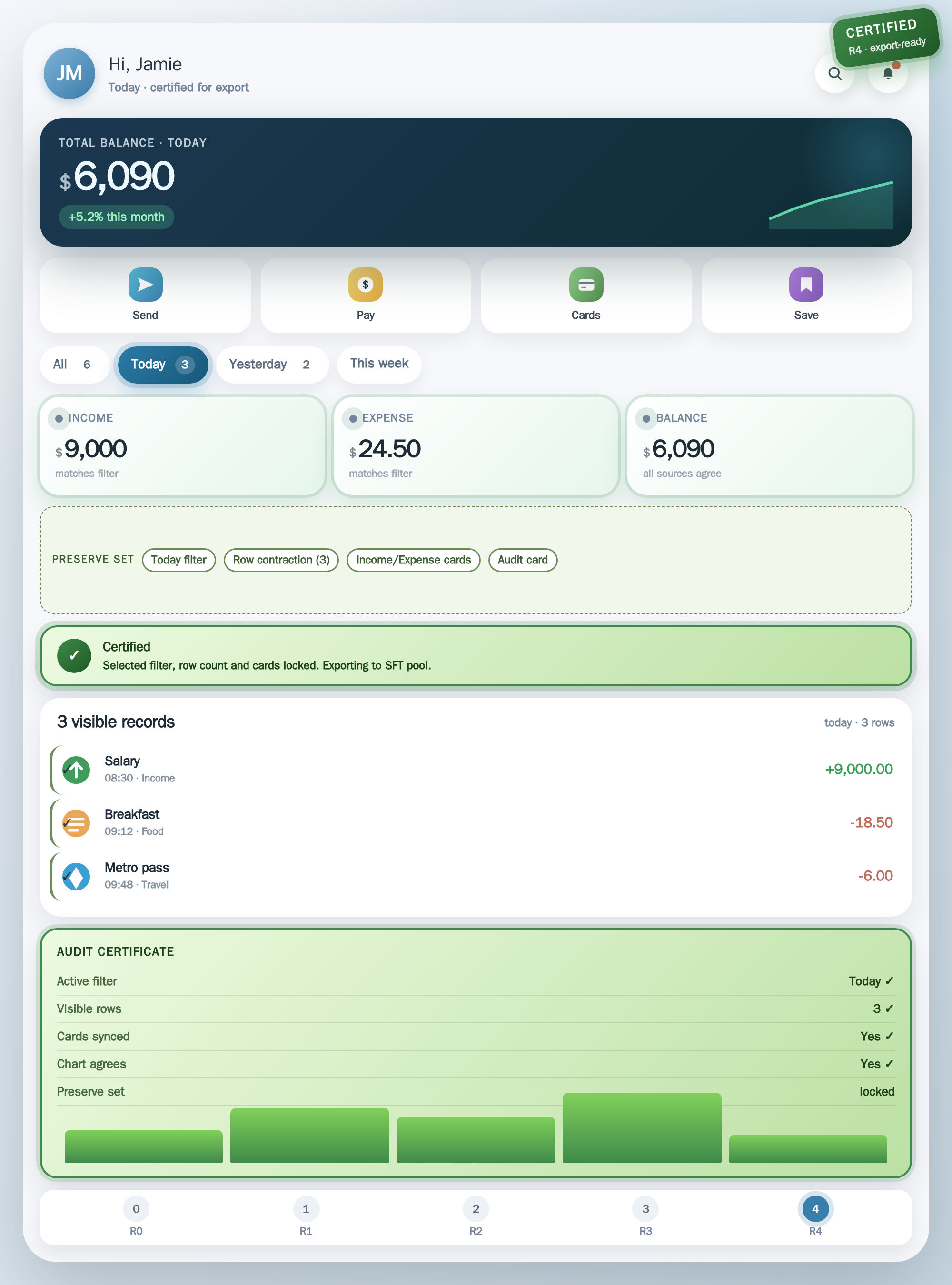}{fig:case_finance_04}{Round 4 accepted: auditable filter state.}
\caseannot{\texttt{target}\,=\,\texttt{audit-card.}\allowbreak\texttt{attached}\,$\land$\,\texttt{filter.}\allowbreak\texttt{exportable};\ \texttt{replay}\,=\,\texttt{full-rerun}\allowbreak\texttt{+probe}\allowbreak\texttt{(filter,rows,cards)};\ \texttt{preserve}(5)\,=\,prev\,$\cup$\,\{\texttt{audit-card.}\allowbreak\texttt{attached},\allowbreak\,\texttt{filter.}\allowbreak\texttt{exportable}\}}{\textsc{export-ready} via static structural proof; the certified trajectory enters the SFT export pool}
\noindent\textbf{Interpretation.}
Round 4 adds an audit card that explicitly names the active filter, row count, and synchronized cards. The figure and explanation are aligned: the final dashboard is more polished because its visual summary is also replay-auditable.

\subsection{Case C: Knife Master Click-State Repair}
\label{app:case_knife_throw}

\caseimage{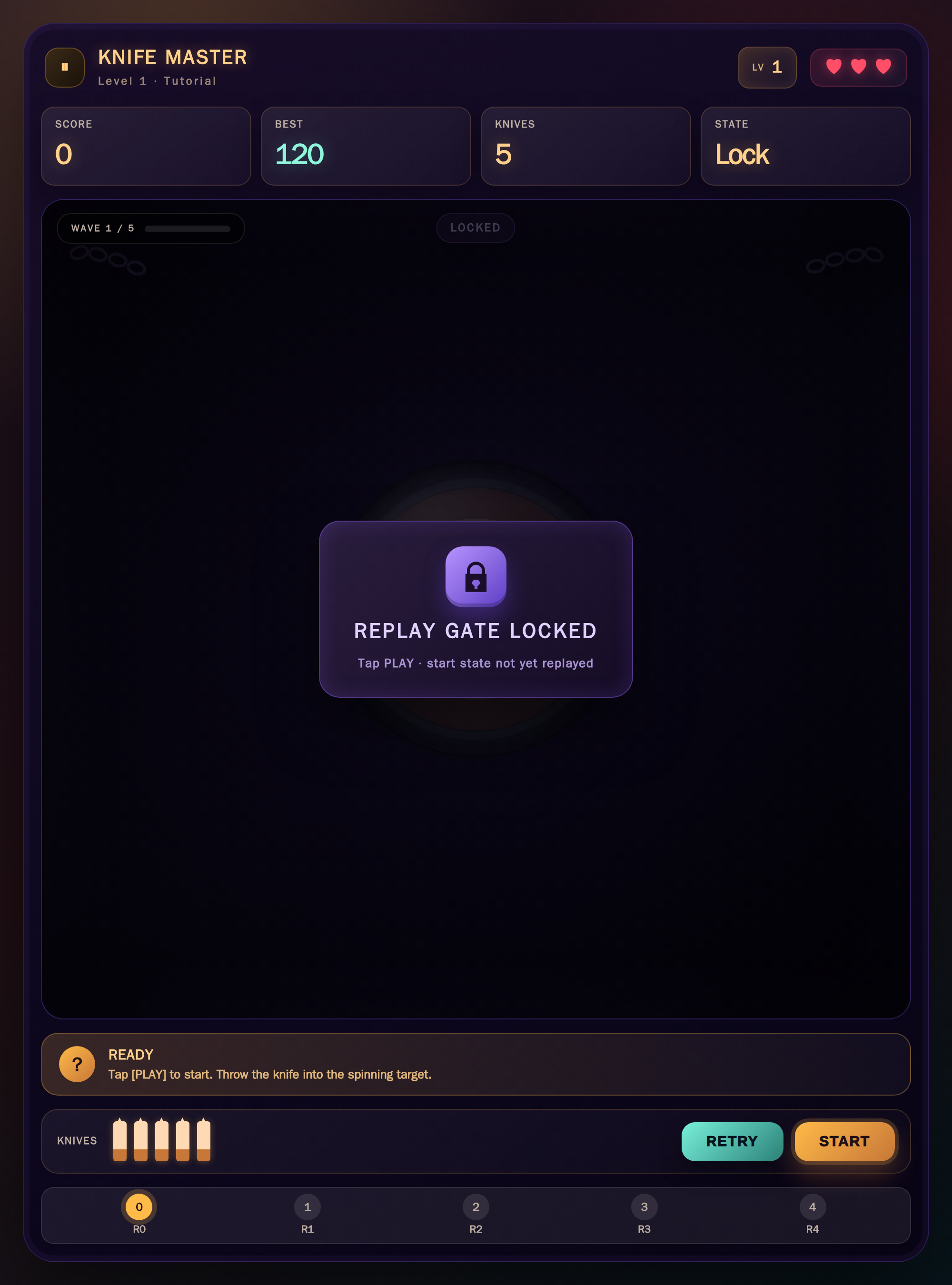}{fig:case_knife_00}{Knife Master initial state.}
\caseannot{(no contract yet --- this is the unverified baseline page)}{no certificate; the page has not been re-executed under any contract}
\noindent\textbf{Interpretation.}
The initial game example separates the target, score tiles, and action buttons, but it remains a baseline because the Start action has not yet been replayed. The right-side status card explicitly marks this unverified state.

\caseimage{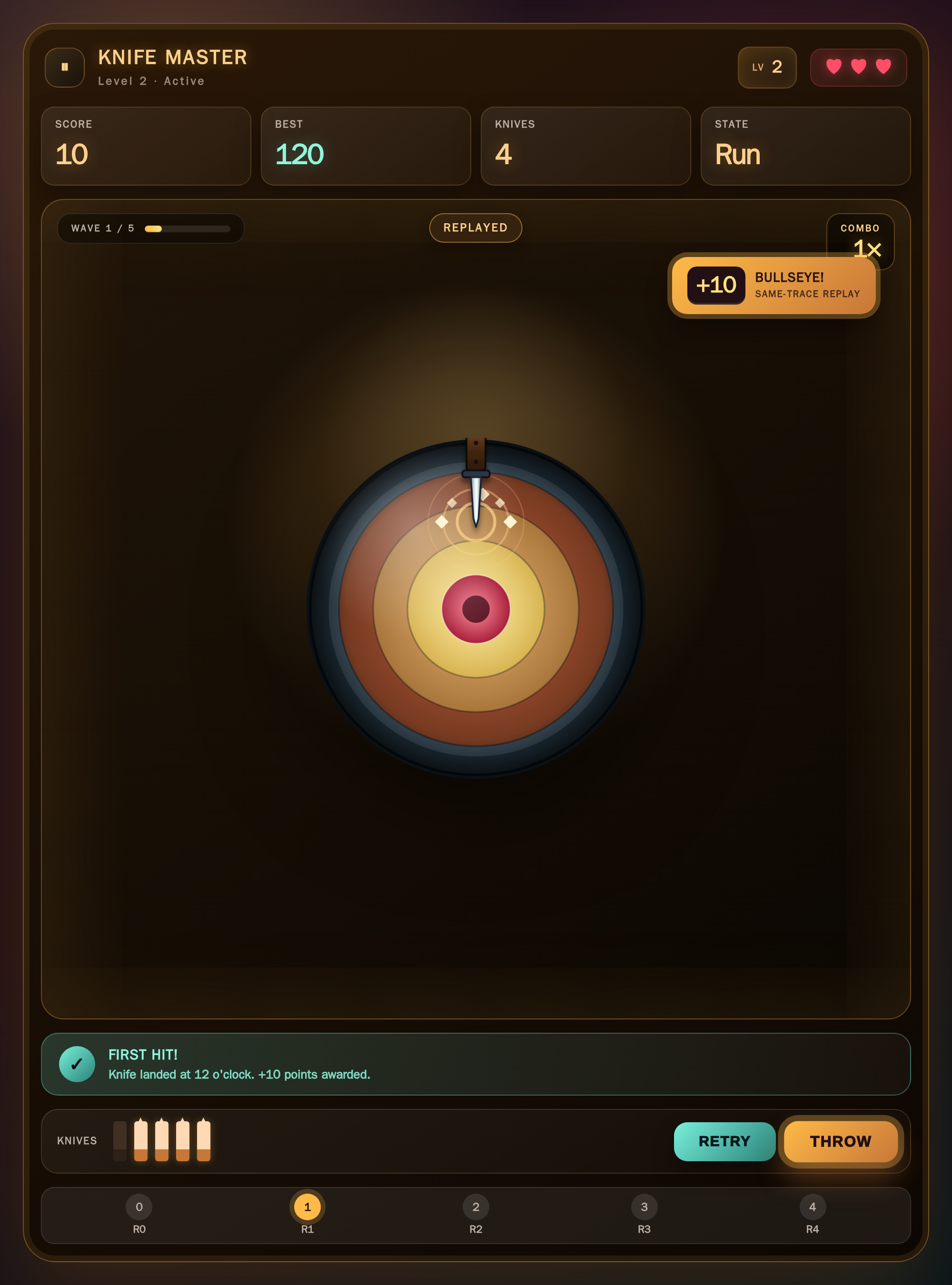}{fig:case_knife_01}{Round 1 accepted: start control.}
\caseannot{\texttt{target}\,=\,\texttt{state\_changed}\allowbreak\texttt{(button.start-game)};\ \texttt{replay}\,=\,\texttt{click(button.start-game)}\allowbreak\texttt{+wait(300ms)}\allowbreak\texttt{+screenshot};\ \texttt{preserve}(1)\,=\,\{\texttt{button.start-game.replayed}\}}{\textsc{accepted} via same-trace replay proof}
\noindent\textbf{Interpretation.}
Round 1 accepts the first interaction repair: the Start button is replayed while the score tiles, target, and action area stay separated. The visual improvement is modest but important: no control is hidden by the target or knives.

\caseimage{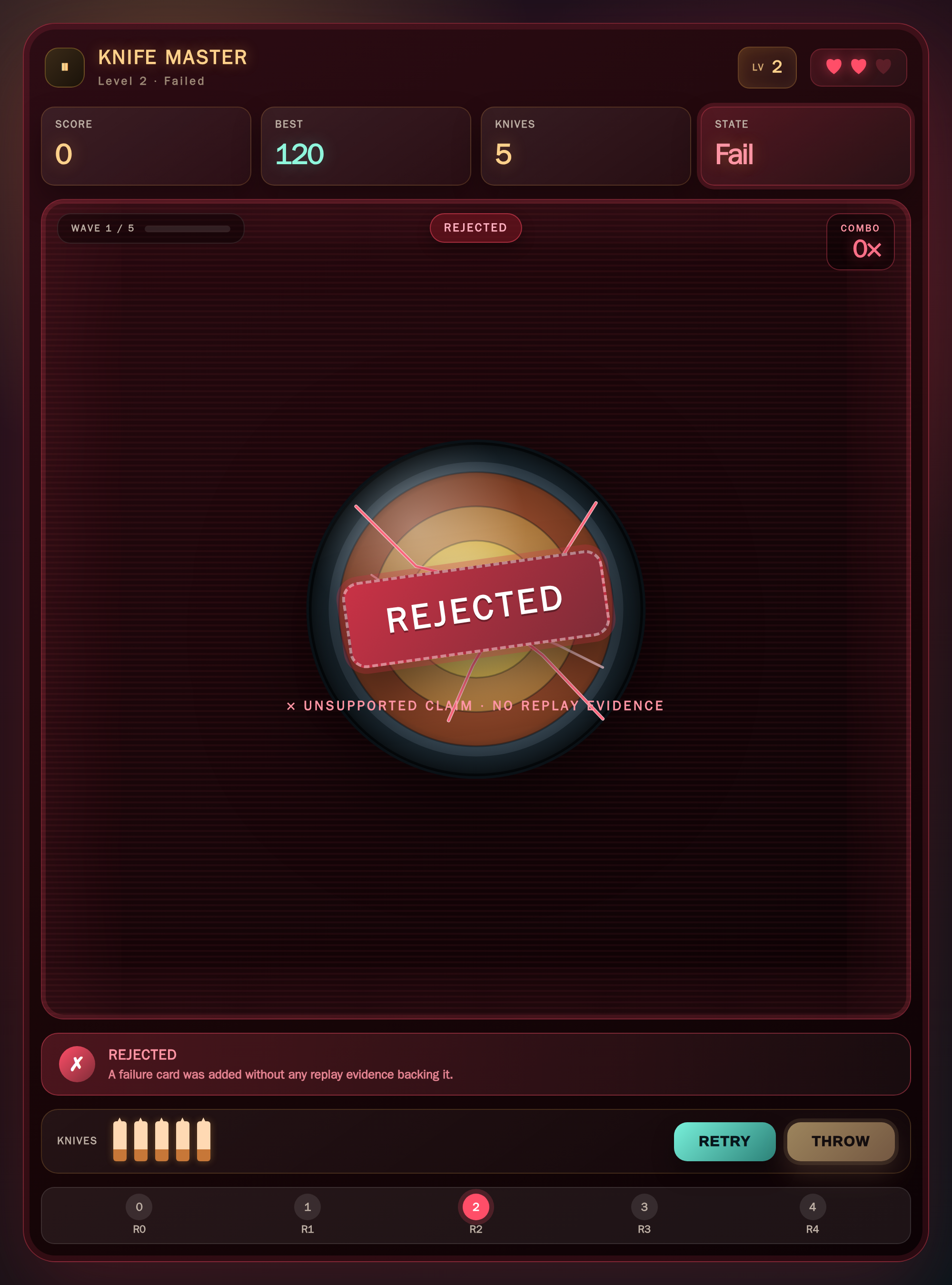}{fig:case_knife_02}{Round 2 rejected: unsupported claim.}
\caseannot{\texttt{target}\,=\,\texttt{failure-card.}\allowbreak\texttt{semantic-evidence};\ \texttt{replay}\,=\,no concrete replay action provided;\ \texttt{preserve}\,=\,(unchanged from R1)}{\textsc{rejected} --- the candidate adds a failure card but provides no replayed evidence backing it, so no certificate is issued}
\noindent\textbf{Interpretation.}
Round 2 is the rejected variant. It adds a prominent failure card and warning, but the candidate claim is not grounded by replay evidence. The red decision panel is synchronized with that visible warning, so the attractive-looking edit is rejected by the gate.

\caseimage{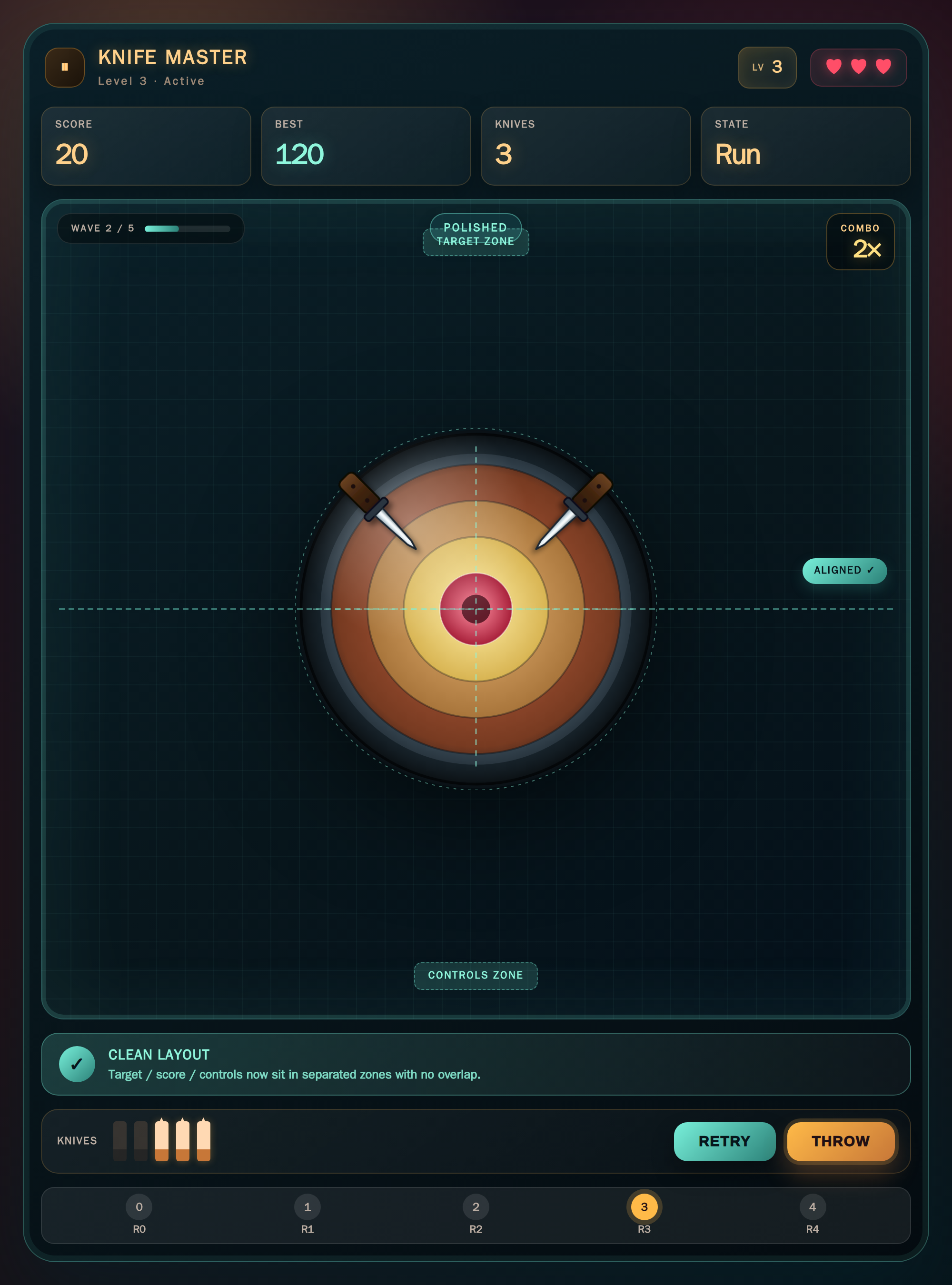}{fig:case_knife_03}{Round 3 accepted: polished layout.}
\caseannot{\texttt{target}\,=\,\texttt{no-overlap}\allowbreak\texttt{(target,controls,score)};\ \texttt{replay}\,=\,\texttt{rerun}\allowbreak\texttt{+visual-cmp(playfield)};\ \texttt{preserve}(3)\,=\,prev\,$\cup$\,\{\texttt{no-overlap},\allowbreak\,\texttt{score-tiles.visible}\}}{\textsc{accepted} via localized visual proof}
\noindent\textbf{Interpretation.}
Round 3 is the accepted aesthetic repair: the target is centered, blades stay inside the play area, and the action buttons move into a separate control deck. The screenshot now visibly matches the explanation: the UI becomes cleaner without losing the replayed state.

\caseimage{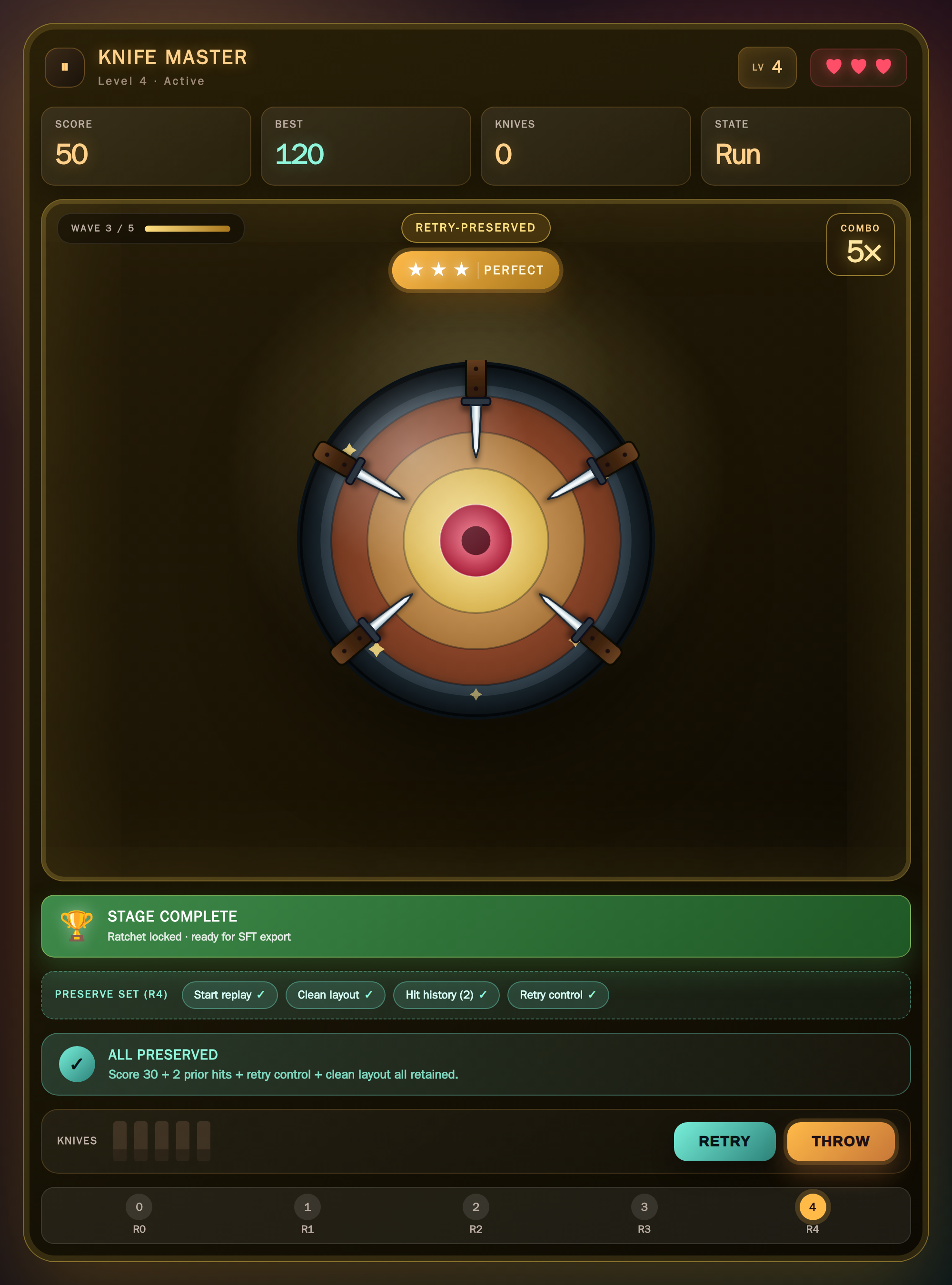}{fig:case_knife_04}{Round 4 accepted: retry state preserved.}
\caseannot{\texttt{target}\,=\,\texttt{retry-card.visible}\,$\land$\,prior-state-holds;\ \texttt{replay}\,=\,\texttt{full-rerun}\allowbreak\texttt{+probe(retry,score,layout)};\ \texttt{preserve}(4)\,=\,prev\,$\cup$\,\{\texttt{retry-card.visible}\}}{\textsc{export-ready} via capability gain proof; the certified trajectory enters the SFT export pool}
\noindent\textbf{Interpretation.}
Round 4 preserves the polished layout while adding an explicit retry/status card. The failure card, retry button, score tiles, target, knives, and main controls are all visible without overlap, which is the case-study point: replay and preservation, not polish alone, decide acceptance.

\clearpage
\section{System Inventory}
\label{app:implementation}

This appendix lists the concrete tools the prototype carries. The method in \S\ref{sec:method} does not depend on this inventory; tool changes preserve the interface.

\paragraph{Interaction probes.}
The runtime drives the rendered page through fourteen deterministic browser macros, each a fixed sequence of freeze, snapshot, action, and observe steps. The catalogue covers four behavioral surfaces: \textbf{rendering and entry} (canvas focus, primary call-to-action, game start), \textbf{control wiring} (click target, button sweep, keyboard, pointer motion, pointer hold), \textbf{drag and form input} (drag, form submit, filter change), and \textbf{interactive visuals} (card flip, hover reveal, SVG animation). Each macro yields replay evidence.

\paragraph{Repair skills.}
The repair router maps issue families to twenty typed skills. Fifteen are VLM lanes grouped as \textbf{gameplay} (game-loop and playability fixes), \textbf{control wiring} (click, drag, form handlers), \textbf{SVG} (animation, card-flip, structural), \textbf{structural completion} (feature completion, interaction affordance), \textbf{visual} (polish, enrichment), \textbf{maintenance} (cleanup, console-error fixes), and a bounded holistic-rewrite fallback. Five deterministic patch lanes bypass the VLM for mechanical fixes: broken console state, missing dependency, missing SVG hook, missing interaction affordance, or missing focus feedback.

\paragraph{Contract schema.}
The serialized interaction contract has a stable version tag and records: the issue family, the target predicate, the action trace and stable selectors the verifier will replay, the inherited preserve predicates, the hard gates (same-trace execution, no new fatal console errors, no regression on previously accepted contracts), an impact-scope region, and an admission record that names which planner rule admitted it.

\paragraph{Certificate schema.}
The serialized acceptance certificate has a stable identifier and five blocks: the restated objective, the HTML-level diff summary, the proof block (proof level, state-trigger evidence, target and summary terms), the validation result per gate and per success criterion, and the provenance bundle (screenshots, action trace, probe outcomes). Refusals share the same identifier scheme but carry a typed rejection in place of the proof block.

\paragraph{Reproducibility note.}
The prototype records both live verification metrics and metadata-derived accepted-artifact metrics. When these disagree, the main claims use the stricter verification-derived accounting; live metrics are runtime diagnostics.

\end{document}

%% file: table_main_results.tex
\begin{table*}[!t]
\centering
\scriptsize
\setlength{\tabcolsep}{2.2pt}
\renewcommand{\arraystretch}{0.99}
\newcommand{\resultgroup}[1]{%
  \arrayrulecolor{black!65}\specialrule{0.055em}{0pt}{0pt}%
  \rowcolor{HeaderColor}
  \multicolumn{13}{c}{\rule[-0.30ex]{0pt}{2.05ex}\smash{\textbf{#1}}}\\[-0.8pt]
  \arrayrulecolor{black!65}\specialrule{0.055em}{0pt}{0pt}%
  \arrayrulecolor{black}%
}
\caption{Main results on HTMLBench and the released MiniAppBench validation split. MoE sizes report total and active parameters when disclosed. SFT routes share backbone family, optimizer, LR schedule, prompt template, and evaluation runner; each SFT cell is a single seed-42 run. Full configuration in \autoref{app:sft_plan}.}
\label{tab:main_sft_results}
\resizebox{0.98\textwidth}{!}{%
\begin{tabular}{llrrrrrrrrrrr}
\toprule
\multirow{2}{*}{\textbf{Model}} &
\multirow{2}{*}{\textbf{Size}} &
\multicolumn{7}{c}{\textbf{HTMLBench-400}} &
\multicolumn{4}{c}{\textbf{MiniAppBench-Val}} \\
\cmidrule(lr){3-9} \cmidrule(lr){10-13}
& & \textbf{Score} & \textbf{TC Pass (\%)} &
\textbf{Rend.} & \textbf{Vis.} & \textbf{TC/Func.} & \textbf{Inter.} & \textbf{Code} &
\textbf{Intent} & \textbf{Static} & \textbf{Dyn.} & \textbf{Avg.} \\
\resultgroup{Open source models}
\rowcolor{GoodColor}
Kimi-K2.6 & 1T${\mathrm{A}}$32B & \textbf{49.8} & \textbf{43.7} & 8.7 & 10.8 & \textbf{24.8} & 1.2 & \textbf{4.1} & 92.2 & 85.6 & 78.9 & 85.5 \\
\rowcolor{GoodColor}
Kimi-K2.5 & 1T${\mathrm{A}}$32B & 49.4 & 42.5 & 9.1 & 11.2 & 24.1 & 1.1 & 4.0 & 89.6 & 82.9 & 76.2 & 82.9 \\
\rowcolor{GoodColor}
GLM-5 & 744B${\mathrm{A}}$40B & 49.2 & 41.9 & 8.9 & 11.2 & 23.7 & 1.3 & 4.0 & 90.9 & 81.0 & 74.2 & 82.0 \\
\rowcolor{GoodColor}
DeepSeek-V3.2 & 685B${\mathrm{A}}$37B & 48.8 & 42.8 & 8.5 & 10.5 & 24.3 & \textbf{1.4} & \textbf{4.1} & 89.8 & 81.2 & 75.1 & 82.0 \\
\rowcolor{GoodColor}
GLM-5.1 & 754B${\mathrm{A}}$40B & 48.6 & 39.6 & 9.2 & \textbf{11.7} & 22.5 & 1.2 & 4.0 & \textbf{92.8} & \textbf{87.0} & \textbf{83.6} & \textbf{87.8} \\
\rowcolor{GoodColor}
GLM-4.7 & 358B${\mathrm{A}}$32B & 47.7 & 40.7 & 8.9 & 10.6 & 23.1 & 1.1 & \textbf{4.1} & 88.8 & 82.2 & 73.0 & 81.3 \\
\rowcolor{GoodColor}
Qwen3.6-27B & 27B & 47.1 & 38.1 & \textbf{9.3} & 11.0 & 21.6 & 1.1 & 4.0 & 91.7 & 84.5 & 76.8 & 84.3 \\
\rowcolor{GoodColor}
Qwen3.6-35B-A3B & 35B${\mathrm{A}}$3B & 46.3 & 37.4 & \textbf{9.3} & 10.8 & 21.2 & 1.0 & 4.0 & 90.0 & 83.1 & 72.2 & 81.8 \\
\rowcolor{GoodColor}
Qwen3.5-122B-A10B & 122B${\mathrm{A}}$10B & 45.6 & 38.0 & 9.0 & 10.0 & 21.5 & 1.0 & \textbf{4.1} & 87.2 & 80.7 & 77.8 & 81.9 \\
\rowcolor{GoodColor}
Qwen3.5-397B-A17B & 397B${\mathrm{A}}$17B & 45.6 & 37.3 & 9.2 & 10.1 & 21.1 & 1.1 & \textbf{4.1} & 84.4 & 77.3 & 61.7 & 74.5 \\
\rowcolor{GoodColor}
Qwen3.5-27B & 27B & 45.5 & 38.4 & 8.7 & 9.8 & 21.8 & 1.1 & 4.0 & 87.6 & 73.7 & 68.3 & 76.5 \\
\rowcolor{GoodColor}
Qwen3.5-35B-A3B & 35B${\mathrm{A}}$3B & 45.2 & 37.7 & 9.0 & 9.8 & 21.3 & 1.0 & \textbf{4.1} & 82.9 & 74.3 & 59.0 & 72.1 \\
\rowcolor{GoodColor}
DeepSeek-V4-Flash & 284B${\mathrm{A}}$13B & 44.5 & 35.7 & 8.5 & 10.7 & 20.3 & 1.1 & 3.9 & 90.1 & 82.0 & 83.1 & 85.1 \\
\rowcolor{GoodColor}
MiniMax-M2.5 & 229B${\mathrm{A}}$10B & 44.5 & 36.9 & 8.9 & 9.8 & 20.9 & 0.9 & 4.0 & 90.0 & 83.2 & 76.3 & 83.2 \\
\rowcolor{GoodColor}
Qwen3.5-9B & 9B & 42.9 & 35.1 & 8.9 & 9.1 & 19.9 & 0.9 & 4.0 & 73.3 & 63.2 & 30.7 & 55.7 \\
\rowcolor{GoodColor}
Qwen3.5-4B & 4B & 41.7 & 35.4 & 8.7 & 8.5 & 19.6 & 0.9 & 4.0 & 67.2 & 57.2 & 20.3 & 48.2 \\
\resultgroup{Closed source models}
\rowcolor{ClosedColor}
GPT-5.4 & \faLock &\textbf{49.2} & \textbf{42.7} & 8.8 & 11.0 & \textbf{24.2} & \textbf{1.0} & \textbf{4.2} & \textbf{90.8} & \textbf{86.5} & 85.6 & \textbf{87.7} \\
\rowcolor{ClosedColor}
Claude-Opus-4.7 & \faLock &47.8 & 39.8 & 9.2 & \textbf{11.1} & 22.6 & \textbf{1.0} & 3.9 & 89.7 & 83.3 & \textbf{85.9} & 86.3 \\
\rowcolor{ClosedColor}
Claude-Opus-4.6 & \faLock &47.5 & 39.3 & 9.2 & 11.0 & 22.3 & 0.9 & 4.1 & 87.3 & 81.8 & 81.9 & 83.7 \\
\rowcolor{ClosedColor}
Claude-Sonnet-4.6 & \faLock &46.6 & 37.2 & \textbf{9.4} & \textbf{11.1} & 21.1 & \textbf{1.0} & 3.9 & 87.9 & 83.4 & 80.8 & 84.0 \\
\rowcolor{ClosedColor}
Gemini-3.1-Pro & \faLock &45.4 & 37.0 & 8.9 & 10.5 & 21.0 & 0.9 & 4.0 & 84.3 & 78.8 & 78.3 & 80.5 \\
\rowcolor{ClosedColor}
Claude-Opus-4.5-20251101 & \faLock &43.9 & 38.5 & 8.8 & 8.9 & 21.8 & \textbf{1.0} & 3.3 & 88.0 & 82.2 & 77.8 & 82.7 \\
\resultgroup{\method SFT models}
\rowcolor{FairColor}
\method-4B-Raw & 4B & 43.3 & 26.8 & 9.9 & 8.1 & 18.9 & 0.5 & 5.9 & 68.2 & 57.9 & 24.7 & 50.3 \\
\rowcolor{OurColor}
\textbf{\method-4B} & \textbf{4B} & \textbf{46.7} & \textbf{39.3} & 9.5 & \textbf{9.9} & 22.3 & \textbf{0.9} & 4.2 & 76.6 & 66.7 & 39.2 & 60.8 \\
\rowcolor{FairColor}
\method-9B-Raw & 9B & 44.5 & 34.1 & 9.5 & 9.1 & 20.3 & 0.7 & 4.9 & 74.1 & 61.6 & 23.0 & 52.9 \\
\rowcolor{OurColor}
\textbf{\method-9B} & \textbf{9B} & \textbf{49.3} & \textbf{40.6} & 9.5 & 8.0 & 26.1 & 0.6 & 5.1 & 81.2 & 71.6 & 46.0 & 66.3 \\
\rowcolor{FairColor}
\method-27B-Raw & 27B & 47.4 & 33.5 & 9.8 & 9.3 & 21.7 & 0.7 & 5.9 & 79.3 & 71.2 & 61.2 & 70.6 \\
\rowcolor{OurColor}
\textbf{\method-27B} & \textbf{27B} & \textbf{52.7} & \textbf{43.2} & 9.5 & 8.9 & \textbf{28.0} & 0.6 & 5.7 & \textbf{92.7} & \textbf{85.4} & \textbf{78.3} & \textbf{85.5} \\
\bottomrule
\end{tabular}}
\vspace{-5pt}
\end{table*}

%% file: acl2023.bbl
\begin{thebibliography}{24}
\expandafter\ifx\csname natexlab\endcsname\relax\def\natexlab#1{#1}\fi

\bibitem[{Chen et~al.(2023)Chen, Lin, Sch{\"a}rli, and
  Zhou}]{chen2023selfdebugging}
Xinyun Chen, Maxwell Lin, Nathanael Sch{\"a}rli, and Denny Zhou. 2023.
\newblock \href {http://arxiv.org/abs/2304.05128} {Teaching large language
  models to self-debug}.

\bibitem[{Deng et~al.(2023)Deng, Gu, Zheng, Chen, Stevens, Wang, Sun, and
  Su}]{deng2023mind2web}
Xiang Deng, Yu~Gu, Boyuan Zheng, Shijie Chen, Samuel Stevens, Boshi Wang, Huan
  Sun, and Yu~Su. 2023.
\newblock \href {http://arxiv.org/abs/2306.06070} {Mind2web: Towards a
  generalist agent for the web}.

\bibitem[{Gunasekar et~al.(2023)Gunasekar, Zhang, Aneja, Mendes, Del~Giorno,
  Gopi, Javaheripi, Kauffmann, de~Rosa, Saarikivi, Salim, Shah, Behl, Wang,
  Bubeck, Eldan, Kalai, Lee, and Li}]{gunasekar2023textbooks}
Suriya Gunasekar, Yi~Zhang, Jyoti Aneja, Caio C{\'e}sar~Teodoro Mendes, Allie
  Del~Giorno, Sivakanth Gopi, Mojan Javaheripi, Piero Kauffmann, Gustavo
  de~Rosa, Olli Saarikivi, Adil Salim, Shital Shah, Harkirat~Singh Behl, Xin
  Wang, S{\'e}bastien Bubeck, Ronen Eldan, Adam~Tauman Kalai, Yin~Tat Lee, and
  Yuanzhi Li. 2023.
\newblock \href {http://arxiv.org/abs/2306.11644} {Textbooks are all you need}.

\bibitem[{He et~al.(2024)He, Yao, Ma, Yu, Dai, Zhang, Lan, and
  Yu}]{he2024webvoyager}
Hongliang He, Wenlin Yao, Kaixin Ma, Wenhao Yu, Yong Dai, Hongming Zhang,
  Zhenzhong Lan, and Dong Yu. 2024.
\newblock \href {https://aclanthology.org/2024.acl-long.371/} {{WebVoyager}:
  Building an end-to-end web agent with large multimodal models}.
\newblock In \emph{Proceedings of the 62nd Annual Meeting of the Association
  for Computational Linguistics}, pages 6864--6890.

\bibitem[{Jimenez et~al.(2024)Jimenez, Yang, Wettig, Yao, Pei, Press, and
  Narasimhan}]{jimenez2024swebench}
Carlos~E. Jimenez, John Yang, Alexander Wettig, Shunyu Yao, Kexin Pei, Ofir
  Press, and Karthik Narasimhan. 2024.
\newblock \href {http://arxiv.org/abs/2310.06770} {Swe-bench: Can language
  models resolve real-world github issues?}

\bibitem[{Lauren{\c{c}}on et~al.(2024)Lauren{\c{c}}on, Tronchon, and
  Sanh}]{laurencon2024websight}
Hugo Lauren{\c{c}}on, L{\'e}o Tronchon, and Victor Sanh. 2024.
\newblock \href {http://arxiv.org/abs/2403.09029} {Unlocking the conversion of
  web screenshots into html code with the websight dataset}.

\bibitem[{Le et~al.(2022)Le, Wang, Gotmare, Savarese, and Hoi}]{le2022coderl}
Hung Le, Yue Wang, Akhilesh~Deepak Gotmare, Silvio Savarese, and Steven
  Chu~Hong Hoi. 2022.
\newblock \href
  {https://proceedings.neurips.cc/paper_files/paper/2022/hash/8636419dea1aa9fbd25fc4248e702da4-Abstract-Conference.html}
  {{CodeRL}: Mastering code generation through pretrained models and deep
  reinforcement learning}.
\newblock In \emph{Advances in Neural Information Processing Systems 35}.

\bibitem[{Liu et~al.(2024)Liu, Song, Lin, Lam, Neubig, Li, and
  Yue}]{liu2024visualwebbench}
Junpeng Liu, Yifan Song, Bill~Yuchen Lin, Wai Lam, Graham Neubig, Yuanzhi Li,
  and Xiang Yue. 2024.
\newblock \href {http://arxiv.org/abs/2404.05955} {Visualwebbench: How far have
  multimodal llms evolved in web page understanding and grounding?}

\bibitem[{Lu et~al.(2025)Lu, Yang, Ren, Hou, Xiao, Wang, Shi, Zhou, Zhan, and
  Li}]{lu2025webgenbench}
Zimu Lu, Yunqiao Yang, Houxing Ren, Haotian Hou, Han Xiao, Ke~Wang, Weikang
  Shi, Aojun Zhou, Mingjie Zhan, and Hongsheng Li. 2025.
\newblock \href {http://arxiv.org/abs/2505.03733} {{WebGen-Bench}: Evaluating
  {LLM}s on generating interactive and functional websites from scratch}.

\bibitem[{Madaan et~al.(2023)Madaan, Tandon, Gupta, Hallinan, Gao, Wiegreffe,
  Alon, Dziri, Prabhumoye, Yang, Gupta, Majumder, Hermann, Welleck,
  Yazdanbakhsh, and Clark}]{madaan2023selfrefine}
Aman Madaan, Niket Tandon, Prakhar Gupta, Skyler Hallinan, Luyu Gao, Sarah
  Wiegreffe, Uri Alon, Nouha Dziri, Shrimai Prabhumoye, Yiming Yang, Shashank
  Gupta, Bodhisattwa~Prasad Majumder, Katherine Hermann, Sean Welleck, Amir
  Yazdanbakhsh, and Peter Clark. 2023.
\newblock \href
  {https://proceedings.neurips.cc/paper_files/paper/2023/hash/91edff07232fb1b55a505a9e9f6c0ff3-Abstract-Conference.html}
  {Self-refine: Iterative refinement with self-feedback}.
\newblock In \emph{Advances in Neural Information Processing Systems 36}.

\bibitem[{Olausson et~al.(2024)Olausson, Inala, Wang, Gao, and
  Solar-Lezama}]{olausson2024selfrepair}
Theo~X. Olausson, Jeevana~Priya Inala, Chenglong Wang, Jianfeng Gao, and
  Armando Solar-Lezama. 2024.
\newblock \href {https://openreview.net/forum?id=y0GJXRungR} {Is self-repair a
  silver bullet for code generation?}
\newblock In \emph{The Twelfth International Conference on Learning
  Representations}.

\bibitem[{Pahuja et~al.(2025)Pahuja, Lu, Rosset, Gou, Mitra, Whitehead, Su, and
  Awadallah}]{pahuja2025explorer}
Vardaan Pahuja, Yadong Lu, Corby Rosset, Boyu Gou, Arindam Mitra, Spencer
  Whitehead, Yu~Su, and Ahmed~Hassan Awadallah. 2025.
\newblock \href {https://aclanthology.org/2025.findings-acl.326/} {Explorer:
  Scaling exploration-driven web trajectory synthesis for multimodal web
  agents}.
\newblock In \emph{Findings of the Association for Computational Linguistics:
  ACL 2025}, pages 6300--6323.

\bibitem[{Roberts et~al.(2024)Roberts, Lee, Wong, Yasunaga, Mai, and
  Liang}]{roberts2024image2struct}
Josselin~S. Roberts, Tony Lee, Chi~H. Wong, Michihiro Yasunaga, Yifan Mai, and
  Percy Liang. 2024.
\newblock \href
  {https://papers.nips.cc/paper_files/paper/2024/hash/d0718553fd6b227a353c6432cf893285-Abstract-Datasets_and_Benchmarks_Track.html}
  {{Image2Struct}: Benchmarking structure extraction for vision-language
  models}.
\newblock In \emph{Advances in Neural Information Processing Systems 37,
  Datasets and Benchmarks Track}.

\bibitem[{Shinn et~al.(2023)Shinn, Cassano, Berman, Gopinath, Narasimhan, and
  Yao}]{shinn2023reflexion}
Noah Shinn, Federico Cassano, Edward Berman, Ashwin Gopinath, Karthik
  Narasimhan, and Shunyu Yao. 2023.
\newblock \href {https://arxiv.org/abs/2303.11366} {Reflexion: Language agents
  with verbal reinforcement learning}.
\newblock In \emph{Advances in Neural Information Processing Systems 36}.

\bibitem[{Si et~al.(2024)Si, Zhang, Li, Yang, Liu, and
  Yang}]{si2024design2code}
Chenglei Si, Yanzhe Zhang, Ryan Li, Zhengyuan Yang, Ruibo Liu, and Diyi Yang.
  2024.
\newblock \href {http://arxiv.org/abs/2403.03163} {Design2code: Benchmarking
  multimodal code generation for automated front-end engineering}.

\bibitem[{Sun et~al.(2025)Sun, Wang, Gu, Li, and Cheng}]{sun2025fullfront}
Haoyu Sun, Huichen~Will Wang, Jiawei Gu, Linjie Li, and Yu~Cheng. 2025.
\newblock \href {http://arxiv.org/abs/2505.17399} {{FullFront}: Benchmarking
  {MLLM}s across the full front-end engineering workflow}.

\bibitem[{Wu et~al.(2026{\natexlab{a}})Wu, Yang, Zhang, Chai, Ma, Shi, Ma, Li,
  and Liu}]{wu-etal-2026-ucoder}
Jiajun Wu, Jian Yang, Wei Zhang, Linzheng Chai, Yuchi Ma, Ensheng Shi, Yuqing
  Ma, Zhoujun Li, and Xianglong Liu. 2026{\natexlab{a}}.
\newblock \href {https://doi.org/10.18653/v1/2026.findings-acl.277} {{UC}oder:
  Unsupervised code generation by internal probing of large language models}.
\newblock In \emph{Findings of the Association for Computational Linguistics:
  ACL 2026}, pages 5642--5655, San Diego, California, United States.
  Association for Computational Linguistics.

\bibitem[{Wu et~al.(2026{\natexlab{b}})Wu, Yang, Zheng, Zhang, Wang, Lou, and
  Liu}]{htmlcure2026}
Jiajun Wu, Jian Yang, Tuney Zheng, Wei Zhang, Haowen Wang, Yihang Lou, and
  Xianglong Liu. 2026{\natexlab{b}}.
\newblock \href {http://arxiv.org/abs/2605.26807} {{HTMLCure}: Turning browser
  experience into state guided repair for interactive {HTML}}.

\bibitem[{Xiao et~al.(2025)Xiao, Wang, Lam, Wan, Liu, Huo, and
  Lyu}]{xiao2025designbench}
Jingyu Xiao, Ming Wang, Man~Ho Lam, Yuxuan Wan, Junliang Liu, Yintong Huo, and
  Michael~R. Lyu. 2025.
\newblock \href {http://arxiv.org/abs/2506.06251} {Designbench: A comprehensive
  benchmark for mllm-based front-end code generation}.

\bibitem[{Yao et~al.(2022)Yao, Chen, Yang, and Narasimhan}]{yao2022webshop}
Shunyu Yao, Howard Chen, John Yang, and Karthik Narasimhan. 2022.
\newblock \href
  {https://proceedings.neurips.cc/paper_files/paper/2022/hash/82ad13ec01f9fe44c01cb91814fd7b8c-Abstract-Conference.html}
  {{WebShop}: Towards scalable real-world web interaction with grounded
  language agents}.
\newblock In \emph{Advances in Neural Information Processing Systems 35}.

\bibitem[{Yao et~al.(2023)Yao, Zhao, Yu, Du, Shafran, Narasimhan, and
  Cao}]{yao2023react}
Shunyu Yao, Jeffrey Zhao, Dian Yu, Nan Du, Izhak Shafran, Karthik~R.
  Narasimhan, and Yuan Cao. 2023.
\newblock \href {https://openreview.net/forum?id=WE_vluYUL-X} {{ReAct}:
  Synergizing reasoning and acting in language models}.
\newblock In \emph{The Eleventh International Conference on Learning
  Representations}.

\bibitem[{Zhang et~al.(2025)Zhang, Li, Xu, Liu, Liu, Zhou, Deng, Wu, Huang, Li,
  Yi, Xiong, Hu, Zhang, Jiang, Xu, Zhang, Zhou, Zhou, and
  Lian}]{zhang2025artifactsbench}
Chenchen Zhang, Yuhang Li, Can Xu, Jiaheng Liu, Ao~Liu, Changzhi Zhou, Ken
  Deng, Dengpeng Wu, Guanhua Huang, Kejiao Li, Qi~Yi, Ruibin Xiong, Shihui Hu,
  Yue Zhang, Yuhao Jiang, Zenan Xu, Yuanxing Zhang, Wiggin Zhou, Chayse Zhou,
  and Fengzong Lian. 2025.
\newblock \href {http://arxiv.org/abs/2507.04952} {Artifactsbench: Bridging the
  visual-interactive gap in llm code generation evaluation}.

\bibitem[{Zhang et~al.(2026)Zhang, Yu, Li, Zhuang, Mo, and
  Li}]{zhang2026miniappbench}
Zuhao Zhang, Chengyue Yu, Yuante Li, Chenyi Zhuang, Linjian Mo, and Shuai Li.
  2026.
\newblock \href {http://arxiv.org/abs/2603.09652} {{MiniAppBench}: Evaluating
  the shift from text to interactive {HTML} responses in {LLM}-powered
  assistants}.

\bibitem[{Zhou et~al.(2024)Zhou, Xu, Zhu, Zhou, Lo, Sridhar, Cheng, Ou, Bisk,
  Fried, Alon, and Neubig}]{zhou2024webarena}
Shuyan Zhou, Frank~F. Xu, Hao Zhu, Xuhui Zhou, Robert Lo, Abishek Sridhar,
  Xianyi Cheng, Tianyue Ou, Yonatan Bisk, Daniel Fried, Uri Alon, and Graham
  Neubig. 2024.
\newblock \href {http://arxiv.org/abs/2307.13854} {Webarena: A realistic web
  environment for building autonomous agents}.

\end{thebibliography}
